\documentclass[sw]{iosart2x}
\makeatletter
\let\numberlines@hook\relax
\makeatother

\usepackage{amsmath,amssymb,enumitem,mathtools}
\usepackage[table,x11names]{xcolor}
\usepackage{soul}

\usepackage[most]{tcolorbox}
\newtcolorbox{highlighted}{colback=yellow,coltext=red,breakable}
\usepackage[htt]{hyphenat}
\usepackage[bordercolor=white,backgroundcolor=gray!30,linecolor=black,colorinlistoftodos]{todonotes}
\usepackage{listings}
\usepackage{graphics}
\usepackage{pdflscape}
\usepackage{afterpage}
\usepackage{capt-of}
\usepackage{pbox}
\usepackage{multirow}
\usepackage{eurosym}
\usepackage{todonotes}
\usepackage{lettrine}
\usepackage[utf8]{inputenc}
\AtBeginDocument{%
  \providecommand\BibTeX{{%
    \normalfont B\kern-0.5em{\scshape i\kern-0.25em b}\kern-0.8em\TeX}}}

\pubyear{0000}
\volume{0}
\firstpage{1}
\lastpage{1}

\begin{document}

\begin{frontmatter}

\title{A Survey on Knowledge Graph Embeddings with Literals:  Which model links better Literal-ly?
}

\runningtitle{Survey on Knowledge Graph Embeddings with Literals}

\author[]{\inits{G. A. G.}\fnms{Genet Asefa} \snm{Gesese}\ead[label=e1]{genet-asefa.gesese@fiz-karlsruhe.de}%
\thanks{Corresponding author. \printead{e1}.}},
\author[]{\inits{R. B.}\fnms{Russa} \snm{Biswas}\ead[label=e2]{	russa.biswas@fiz-karlsruhe.de}},
\author[]{\inits{M. A.}\fnms{Mehwish} \snm{Alam}\ead[label=e3]{mehwish.alam@fiz-karlsruhe.de}},
and
\author[]{\inits{H. S.}\fnms{Harald} \snm{Sack}\ead[label=e4]{harald.sack@fiz-karlsruhe.de}}
\runningauthor{G. A. Gesese et al.}
\address{\orgname{FIZ Karlsruhe - Leibniz Institute for Information Infrastructure \& Institute for Applied Informatics and Formal Description Systems (AIFB), Karlsruhe Institute of Technology}, Karlsruhe \cny{Germany}\printead[presep={\\}]{e1,e2,e3,e4}}

\begin{abstract}
Knowledge Graphs (KGs) are composed of structured information about a particular domain in the form of entities and relations. In addition to the structured information KGs help in facilitating interconnectivity and interoperability between different resources represented in the Linked Data Cloud. KGs have been used in a variety of applications such as entity linking, question answering, recommender systems, etc.
However, KG applications suffer from high computational and storage costs. Hence, there arises the necessity for a representation able to map the high dimensional KGs into low dimensional spaces, i.e., embedding space, preserving structural as well as relational information. This paper conducts a survey of KG embedding models which not only consider the structured information contained in the form of entities and relations in a KG but also the unstructured information represented as literals such as text, numerical values, images, etc. Along with a theoretical analysis and comparison of the methods proposed so far for generating KG embeddings with literals, an empirical evaluation of the different methods under identical settings has been performed for the general task of link prediction. 

\end{abstract} 


\end{frontmatter}


\section{Introduction}
\label{sec:introduction}

\lettrine[findent=2pt]{\textbf{K}}{ }nowledge Graphs (KGs) have become quite crucial for storing structured information. There has been a sudden attention towards using KGs for various applications mainly in the area of artificial intelligence. For instance, in a more general sense, KGs can be used to support decision making process and to improve different machine learning applications such as question answering \cite{bordes-etal-2014-question}, recommender systems \cite{CollaborativeRecommenderSystem}, and relation extraction \cite{relationExtraction}. Some of the most popular publicly available general purpose KGs are DBpedia \cite{lehmann2015dbpedia}, Wikidata \cite{vrandevcic2014wikidata}, and YAGO \cite{mahdisoltani2013yago3}. These general purpose KGs often consist of huge amount of facts constructed using 
millions 
of entities (represented as nodes) and relations (as edges connecting these nodes). 
 
Although KGs are effective in representing structured data, there exist some issues which hinder their efficient manipulation such as i) different KGs are usually based on different rigorous symbolic frameworks and this makes it hard to utilize their data in other applications \cite{LearningStruEmbeddings} and ii) the fact that a significant number of important graph algorithms needed for the efficient manipulation and analysis of graphs have proven to be NP-complete \cite{Garey:1990:CIG:574848}. In order to address these issues and use a KG more efficiently, it is beneficial to transform it into a low dimensional vector space while preserving its underlying semantics. To this end, various attempts have been made so far to learn vector representations (embeddings) for KGs. 

As discussed in {\cite{wang2017Know}}, a typical KG embedding approach, which uses only structured information from the KG, generally follows three steps: (i) determining the form of entity and relation representations, (ii) defining a scoring function, and (iii) learning entity and relation representations. In the first step, the forms in which entities and relations are represented in the vector space are determined. Entities can be represented as vectors or modeled as multivariate Gaussian distributions whereas relations can be encoded as operations, matrices, tensors, multivariate Gaussian distributions, or mixtures of Gaussians. Once the form of the entities are determined, in the second step, a scoring function which measures the plausibility of a triple is defined. The main goal is to enable the model to assign higher score to true triples and lower scores to false/negative/corrupted triples. Thus, in order to achieve this, the third step solves an optimization problem which maximizes the plausibility of true facts in order to learn the embeddings of entities and relations. Note that the method used to generate false/negative/corrupted triples has an impact on the performance of a model. The various negative triple generation methods and their differences are discussed in detail in {\cite{kotnis2017analysis}}.

Among the different embedding approaches proposed so far, TransE {\cite{bordes2013translating}} is, to the best of our knowledge, the very first attempt to use distance-based scoring function to learn KG embedding. Given a triple $<h,r,t>$ where $h$ and $t$ are head and tail entities respectively and $r$ is a relation, TransE represents $h$, $r$, and $t$ as vectors $\textbf{h}$, $\textbf{r}$, and $\textbf{t}$ respectively by modeling the relation $r$ as a translation vector which connects the vectors $\textbf{h}$ and $\textbf{t}$. The problem with TransE is that it fails to model certain type of relations such as one-to-many or many-to-one. In order to address such limitations, different embedding techniques which are extension of TransE or are entirely new have been proposed. However, most of the existing approaches, including the current state-of-the-art models such as ConvE {\cite{dettmers2018convolutional}}, are structure-based embeddings which do not make use of any literal information i.e., only triples consisting of entities connected via properties are usually considered. 
This is a major disadvantage because information encoded in the literals will be left unused when capturing the semantics of a certain entity. 

Literals can bring advantages to the process of learning KG embeddings in two major ways:
\begin{enumerate}
    \item \textit{Learning embeddings for novel entities:} Novel entities are the entities which are not linked to any other entity in the KG but have literal values associated with them such as 
    their \textit{textual description, numeric literals, and images}. 
    In most existing structure-based embedding models, it is not possible to learn embeddings for such novel entities. However, this can be addressed by utilizing the information represented in literals to learn embeddings. For example, considering the dataset FB15K-20 \cite{xie2016representation}, which is a subset of Freebase, the entity \texttt{'/m/0gjd61t'} is a novel entity which does not occur in any of the training triples, but it has a description given as follows in the form \textit{<subject, relation, object>}.
  
    \begin{lstlisting}[basicstyle=\footnotesize\ttfamily]
    </m/0gjd61t, http://rdf.freebase.com/ns/common.topic.description, "Vincent Franklin is an English actor best known for his roles in comedy television programmes...">
    \end{lstlisting}
    In order to learn the embedding for this particular entity (i.e., \texttt{/m/0gjd61t}),
    the model can make use of the entity's textual description. 
    DKRL \cite{xie2016representation} is one of those approaches which provide embeddings for novel entities using their descriptions.

    \item \textit{Improving the representation of entities in structure based embedding models:} Literals play a vital role in improving the representation learning where an entity is required to appear in at least a minimum number of relational triples. For example, taking into consideration only the information provided in a sample KG presented in Figure~\ref{fig:kg}, which is extracted from DBpedia, it is not possible to tell apart the entities 
    \texttt{dbr:Gina\_Torres}, \texttt{dbr:Patrick\_J.Adams}, and \texttt{dbr:Sarah\_Rafferty} 
    from one another. This is the case due to the fact that the only information that is available regarding these entities in this KG is that they all star in the series \texttt{dbr:Suits\_(season\_1)} and this is not enough to know which entities are similar to each other and which are not. Therefore, if some KG embedding model is trained using only this KG, it is not possible to get good representations for the entities 
    \texttt{dbr:Gina\_Torres}, \texttt{dbr:Patrick\_J.Adams}, and \texttt{dbr:Sarah\_Rafferty}. 

 \begin{figure}[h!]
  \centering
  {\includegraphics[width=0.45\textwidth]{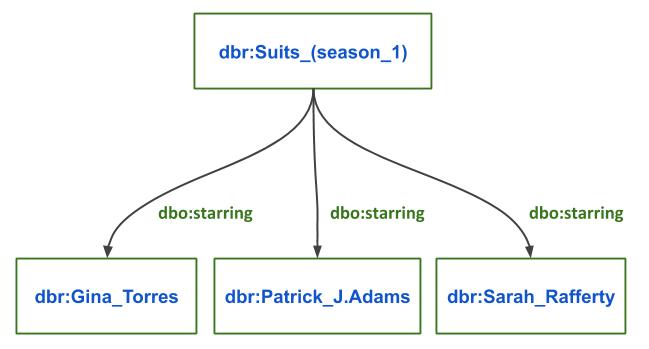}}
    \caption{
    A small fraction of triples taken from the KG DBpedia {\cite{lehmann2015dbpedia}}. }
    \label{fig:kg}
\end{figure}

    However, having the model trained with more triples containing literal values for these entities, as shown in Figure \ref{fig:kg-literal}, would improve the embeddings for the entities. 
    For instance, based on the values of the data relation
    \texttt{dbr:birthDate}, it is possible to deduce the fact that \texttt{dbr:Sarah\_Rafferty} and \texttt{dbr:Gina\_Torres} are almost the same age but they are both older than \texttt{dbr:Patrick\_J.Adams}. On the other hand, the images of the entities along with the textual descriptions (\texttt{dbo:abstract}) would allow us to infer the entities' gender, i.e., \texttt{dbr:Sarah\_Rafferty} and \texttt{dbr:Gina\_Torres} are female and \texttt{dbr:Patrick\_J.Adams} is male.
 
\begin{figure*}
  \centering
  {\includegraphics[width=0.7\textwidth]{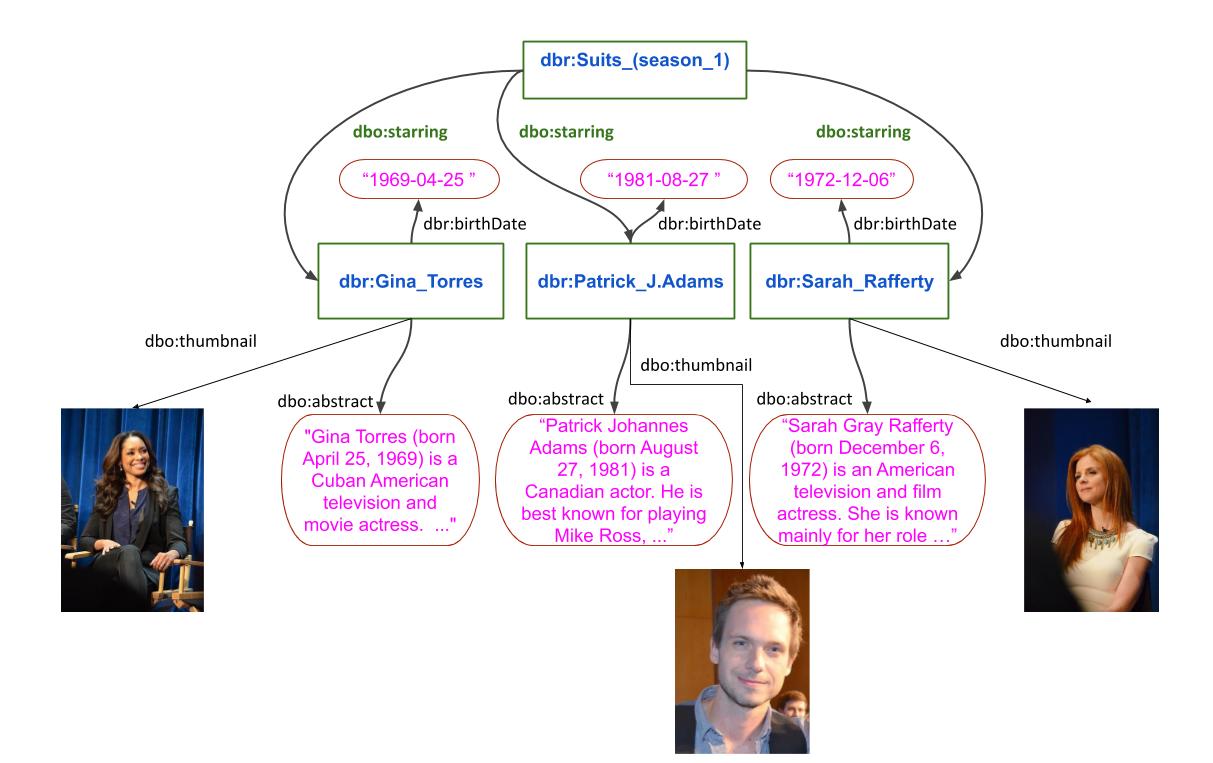}}
  \caption{
  A small fraction of triples with literals taken from the KG DBpedia {\cite{lehmann2015dbpedia}}.}
  \label{fig:kg-literal}
\end{figure*}

     The above example indicates that the use of literals along with their respective entities would add more semantics so that similar entities can be represented close to each other in the vector space while those dissimilar are further apart. 

\end{enumerate} 
Recently, 
some approaches have been proposed which leverage the information present in literals to learn KG embeddings. 
The types of literals considered in these embedding methods are either text, numeric, images, or multi-modal literals, i.e., a combination of more than one medium of information. These methods use different techniques in order to incorporate the literals into the KG embeddings. However, data typed literals are not addressed in these KG embedding models and surveys that are conducted on KG embeddings. The main challenge with data typed literals, such as date and time, is that they require additional semantics to be represented in KG embeddings.

This survey analyses different embedding approaches, which make use of literals, and highlights their advantages and drawbacks in handling different challenges 
such as multi-valued data properties/relations, data typed literals, and units of literals. 
A review of the different applications used for model evaluation by different KG embedding models is also presented. 
Furthermore, experiments with some of the models have been conducted specifically on the link prediction task. 
The contribution of this paper is summarized as follows:
\begin{enumerate}
    \item A detailed analysis of the existing literal enriched KG embedding models and their approaches. In addition, the models are categorized into different classes based on the type of literals used.
    \item An evaluation oriented comparison of the existing models on the link prediction task is performed under same experimental settings.
    \item The research gaps in the area of KG embeddings in using literals are indicated which can open directions for further research.
\end{enumerate}

The rest of this paper is organized as follows: Section~\ref{sec:related_work} presents a brief overview of related work. In Section~\ref{sec:problem_statement}, the problem formulation including definitions, preliminaries, types of literals and research questions are provided while Section~\ref{sec:techniques} analyses different KG embedding techniques with literals is discussed. Section~\ref{sec:application} reviews different tasks used to train or evaluate the embedding models is given. Section~\ref{sec:experimentation} discusses the experiment conducted with the existing KG embedding models with literals on the link prediction task.  
Finally, concluding remarks summarize our findings on KGs with literals and are presented along with future directions in Section~\ref{sec:discussion}. 
 
\section{Related Work}
\label{sec:related_work}

This section describes the state-of-the-art algorithms proposed for generating KG embeddings. It also gives a brief overview of the surveys already published following these lines and what is lacking in those studies. 

A brief overview of the most popular KG embedding techniques, including the state-of-the-art approaches are short listed in Table~{\ref{table:embeddings}}. The categories presented in this table are inspired by a previous survey work {\cite{wang2017Know}} for the models without literals (column 1). The categories are translation based models, semantic matching models, models incorporating entity types, models incorporating relation paths, models using logical rules, models with temporal information, and models using graph structures. We have also categorized the techniques which use literals with respect to the same set of categories.  Since a detailed discussion on these categories on the models without literals has already been presented in {\cite{wang2017Know}}, in the current study, the main focus lies on analysing the models which make use of literals. The standard KG embedding techniques which are extended by the models with literals are listed in Table~{\ref{table:extensions}}. 
 
\begin{table*}[ht!]
    \centering
    \caption{KG embedding models and their categories.} 
    \begin{tabular}{p{4.2cm}|p{6cm}|p{4cm}}
    \hline
    \textbf{Categories} &  \textbf{Models without literals} & \textbf{Models with Literals}\\
    \hline
    Translational Distance Models & TransE \cite{bordes2013translating} and its extensions: TransH \cite{TransH} TransR \cite{TransR}, TransD \cite{TransD}, TranSparse \cite{TranSparse}, TransA \cite{TransA} etc. & 
    TransEA {\cite{Wu2018KnowledgeGE}}, DKRL {\cite{xie2016representation}}, IKRL {\cite{Xie2017ImageembodiedKR}}, Jointly(desp) {\cite{zhong2015aligning}}, Jointly {\cite{ijcai2017xu}}, SSP {\cite{xiao2017ssp}}, KDCoE {\cite{chen2018co}}, EAKGAE {\cite{Trsedya2019entity}}
    \\
    \hline
    \vspace{1mm}
    Semantic Matching Models & 
    RESCAL \cite{nickel2011three} and Its Extensions: DistMult \cite{DistMult}, HolE \cite{HolE}, ComplEx \cite{ComplEx}, and etc.
    Semantic Matching with Neural Networks: SME \cite{SME}, NTN \cite{NTN}, MLP \cite{MLP}, and etc. & 
    LiteralE {\cite{LiteralE2019}}, MKBE {\cite{mmkb:emnlp18}}, MTKGNN {\cite{DBLP:journals/corr/abs-1708-0482}}, KGlove with literals {\cite{cochez2018first}}, Extended RESCAL {\cite{nickel2012factorizing}},  LiteralE with blocking {\cite{de2018towards}} 
    \\ 
    \hline \vspace{1mm}
    Models using Entity Types & SSE \cite{SSE}, TKRL \cite{TKRL}, Type constrained representation learning \cite{TypeConstrained}, Rules incorporated KG completion models \cite{embeedingRules},  TRESCAL \cite{TRESCAL}, Entity Hierarchy Embedding \cite{entityHierarchy} & 
    Extended RESCAL {\cite{nickel2012factorizing}}
    \\
    \hline \vspace{1mm}
    Models using Relation Paths & PTransE \cite{PTransE}, Traversing KGs in Vector Space \cite{guu-etal-2015-traversing}, RTRANSE \cite{garcia-duran-etal-2015-composing}, Compositional vector space \cite{compositionalKBcomp}, Reasoning using RNN \cite{reasoning}, Context-dependent KG embedding \cite{luo-etal-2015-context} & 
    KBLRN {\cite{GarcaDurn2018KBlrnEL}}
    \\
    \hline \vspace{1mm}
    Models using Logical Rules & Rules incorporated KG completion models \cite{embeedingRules}, Large-scale Knowledge Base Completion \cite{Wei2015LargescaleKB}, KALE \cite{KALE}, Logical Background Knowledge for Relation Extraction \cite{rocktaschel-etal-2015-injecting}, and etc. & \\
    \hline 
    Models using Temporal Information & Time-Aware Link Prediction \cite{jiang-etal-2016-encoding}, co-evolution of event and KGs \cite{co-evolution}, Know-evolve \cite{Know-evolve} & \\
    \hline
    Models using Graph Structures & GAKE \cite{GAKE}, Link Prediction in Multi-relational Graphs \cite{LPinMulti-relationalGraphs} & 
    KBLRN {\cite{GarcaDurn2018KBlrnEL}}
    \\
    \hline
    \end{tabular}
    \label{table:embeddings}
\end{table*}

\begin{table*}[ht!]
    \centering
    \caption{
    KG embedding models with literals and their corresponding base models.}
    \begin{tabular}{p{4cm}|p{4cm}}
    \hline
    \textbf{Models with literals} &  \textbf{The standard models they extend} \\
    \hline
    Extended RESCAL {\cite{nickel2012factorizing}} & RESCAL \cite{nickel2011three} \\ 
    Jointly(desp) {\cite{zhong2015aligning}} & TransE \cite{bordes2013translating}\\ 
    DKRL {\cite{xie2016representation}} & TransE \cite{bordes2013translating}\\ 
    Jointly {\cite{ijcai2017xu}} & TransE \cite{bordes2013translating}\\
    SSP {\cite{xiao2017ssp}} & TransE \cite{bordes2013translating}\\
    KDCoE {\cite{chen2018co}} & TransE \cite{bordes2013translating}\\
    KGlove with literals {\cite{cochez2018first}} & KGlove\\
    LiteralE {\cite{LiteralE2019}} & DistMult \cite{DistMult}, ComplEx \cite{ComplEx}, ConvE \cite{dettmers2018convolutional}\\
    TransEA {\cite{Wu2018KnowledgeGE}} & TransE \cite{bordes2013translating} \\
    IKRL {\cite{Xie2017ImageembodiedKR}} & TransE \cite{bordes2013translating}\\
    MTKGRL {\cite{mousselly2018multimodal}} & TransE \cite{bordes2013translating} \\ 
    EAKGAE {\cite{Trsedya2019entity}} & TransE \cite{bordes2013translating}\\
    MKBE {\cite{mmkb:emnlp18}} & DistMult \cite{DistMult}, ConvE \cite{dettmers2018convolutional}\\
    \hline
    \end{tabular}
    \label{table:extensions}
\end{table*}
Few attempts have been made to conduct surveys on the techniques and applications of KG embeddings~\cite{Goyal2018GraphET,Cai2018ACS,wang2017Know}. The survey~\cite{Goyal2018GraphET} is conducted on factorization based, random walk based, and deep learning based network embedding approaches such as DeepWalk, Node2vec, and etc. \cite{Cai2018ACS, wang2017Know} discuss only RESCAL~\cite{nickel2011three} and KREAR~\cite{lin2016knowledge} as methods which use attributes of entities for KG embeddings, and focus mostly on the structure-based embedding methods, i.e., methods using non-attributive triples, for example, translation based embedding models listed in Table~\ref{table:embeddings}. However, RESCAL is a matrix-factorization method for relational learning which encodes each object/data property as a slice of the tensor leading to an increase in the dimensionality of the tensor automatically. This method suffers  from efficiency issues if literals are utilized while generating KG embeddings. Similarly, KREAR only considers those data properties which have categorical values, i.e., fixed number of values and ignores those which take any random literals  as values. One of the recent surveys~\cite{Paulheim17} summarizes the methods proposed so far on refining KGs. However, this survey does not confine itself to embedding techniques and also does not consider most of the approaches which are making use of literals. 
Another very recent related study {\cite{Ruffinelli2020You}}, discusses different aspects of KG embedding models such as model architectures, training strategies, and hyperparameter optimization but it takes into consideration only those models without literals. 

None of the surveys mentioned above include all the existing KG embedding models which make use of literals, such as the ones categorized as models incorporating information represented in literals in Table~\ref{table:embeddings}. To the best of our knowledge, this is the first attempt to analyse the algorithms proposed so far for generating KG embeddings using literals. In this paper, discussions on the type of literals, the embedding approaches, and the applications/tasks on which the embedding models are evaluated are given. A categorization of the models based on the type of literals they use is also provided. 

This survey is an extension of an already published short survey~\cite{GeseseBS19}. The major difference between the two versions is that (i) this survey contains a much more detailed theoretical analysis of the KG embedding models with literals proposed so far, and (ii) it performs empirical evaluation of the discussed models under the same experimental settings under the example of link prediction.  
\section{Problem Formulation}
\label{sec:problem_statement}

This section briefly introduces the fundamentals of KGs and KG embeddings followed by a formal definition of KG embeddings with literals. It also poses various research questions about why conducting this study is a stepping stone for future development.

\subsection{Preliminaries}
\paragraph{\textbf{Knowledge Graphs.}} 
Knowledge Graphs (KGs) consist of a set of triples $K \subseteq E \times R \times (E \cup L)$, where $E$ is a set of resources referred to as entities, $L$ a set of literals, and $R$ a set of relations. An entity is identified by a URI which represents a real-world object or an abstract concept. A relation (or property) is a binary predicate and a literal is a string, date, or number eventually followed by its data type. For a triple \texttt{<h, r, t>}, $h$ is a subject, $r$ is a relation and $t$ is an object. The subject and object are often referred to as \emph{head} and \emph{tail} entity respectively. The triples consisting of literals as objects are often referred to as \emph{attributive triples}. 

\paragraph{\textbf{Relations (or Properties).}}

Based on the nature of the objects, relations are classified into two main categories: 
\begin{itemize}
\item \textbf{Object Relation} links an entity to another entity. E.g., in the triple \\ \texttt{<dbr:Albert\_Einstein, dbo:field, dbr:Physics>}, both \texttt{dbr:Albert\_Einstein} and \texttt{dbr:Physics} are entities, the relation \texttt{dbo:field} is an \textit{Object Relation}.

\item \textbf{Data Type Relation} links an entity to its values, i.e., literals. For example, in \\ \texttt{<dbr:Albert\_Einstein, dbo:birthDate, "1879-03-14">}, where  \texttt{"1879-03-14"} is a literal value, the relation \texttt{dbo:birthDate} is a \textit{Data Type Relation}. 
\end{itemize}

\subsection{Types of Literals} 
Literals in a KG encode additional information which is not captured by the entities or relations. There are different types of literals present in the KGs:

\begin{itemize}

\item \textbf{Text Literals:} A wide variety of information can be stored in KGs in the form of free text such as names, labels, titles, descriptions, comments, etc. In most of the KG embedding models with literals, text information is further categorized into \textbf{\textit{Short text}} and \textbf{\textit{Long text}}. The literals which are fairly short such as for relation like names, titles, labels, etc. are considered as \textit{Short text}. On the other hand, for strings that are much longer such as descriptions of entities, comments, etc. are considered as \textit{Long text} and are usually provided in natural language.

\item \textbf{Numeric Literals:} Information encoded as integers, float and so on such as height, date, population, etc. also provide useful information about an entity. It is worth considering the numbers as distinct entities in the embedding models, as it has its own semantics to be covered which cannot be covered by string distance metrics. For instance, 777 is more similar to 788 than 77.

\item \textbf{Units of Measurement:} Numeric literals often denote units of measurements to a definite magnitude. For example, Wikidata property \texttt{wdt:P2048} 
("height") 
takes values in mm, cm, m, km, inch, foot and pixel. Hence, discarding the units and considering only the numeric values without normalization results in loss of semantics, especially if units are not comparable, e.g., units of length and units of weight.

\item \textbf{Image Literals:} Images also provide latent useful information for modelling the entities. For example, a person's details such as age, gender, etc. can be deduced via visual analysis of an image depicting the person.

\item \textbf{Other Types of Literals:} Useful information encoded in the form of other literals such as external URIs which could lead to an image, text, audio or video files. 
\end{itemize}

\subsection{Research Questions}

As it can be seen from the above discussion that the information represented in the KGs is diverse, modelling these entities is a challenging task. The challenges which are further targeted in this study are given as follows:

\begin{itemize}
\item \textbf{RQ1} -- \textit{How can structured (triples with object relations) and unstructured information (attributive triples) in the KGs be combined into the representation learning?}
\item \textbf{RQ2} -- \textit{How can the heterogeneity of the types of literals present in the KGs be captured and combined into representation learning?} 
\end{itemize}

\section{Knowledge Graph Embeddings with Literals}
\label{sec:techniques}

This section investigates KG embedding models with literals divided into the following different categories based on the types of literals utilized: (i) Text, (ii) Numeric, (iii) Image, and (iv) Multi-modal. A KG embedding model which makes use of at least two types of literals providing complementary information is considered as multi-modal. In the subsequent sections, a description of the models for each of the previously described categories analyzing their similarities and differences, followed by a discussion of potential drawbacks are provided.

\subsection{Models with Text Literals}\label{text}

In this section, seven KG embedding models utilizing text literals are discussed, namely, Extended RESCAL \cite{nickel2012factorizing}, 
Jointly(desp) {\cite{zhong2015aligning}} 
, DKRL \cite{xie2016representation}, 
Jointly {\cite{ijcai2017xu}}, SSP {\cite{xiao2017ssp}} 
, KDCoE \cite{chen2018co}, and KGloVe with literals \cite{cochez2018first}. A detailed description followed by a summary presenting the comparison of these models is given along with their drawback. 
Moreover, in order to show the differences between the models based on complexity, the number of parameters of each model is presented in Table {\ref{tab:complexity_text}}.

\paragraph{{\bf Extended RESCAL}} aims to improve the original RESCAL approach by extending its algorithm to process literal values more efficiently and to deal with the drawback of sparsity that accompanies tensors. In the original RESCAL approach, relational data is modeled as a three-way tensor X of size $n \times n \times m$, where $n$ is the number of entities and $m$ is the number of relations.  An entry $X_{ijk} = 1$ denotes the existence of the triple with i-th entity as a subject, k-th relation as a predicate, and j-th entity as an object. If $X_{ijk}$ is set to 0, it indicates that the triple doesn't exist. A new approach for tensor factorization is proposed which is performed on X. For further details refer to \cite{nickel2012factorizing}. If attributive triples have to be modeled in such a way, then the literals will be taken as entities even if they cannot occur as subject in the triples.  Including literals may lead to an increment in the runtime since a larger tensor has to be factorized. 

In contrast to the original algorithm, the extended RESCAL algorithm handles the attributive triples in a separate matrix. The matrix factorization is performed jointly with the tensor factorization of the non-attributive triples. The attributive triples containing only text literals are encoded in an entity-attribute matrix $D$ in such a way that the rows are entities and the columns are $<data\ type\ relation, value>$ pairs. Given a triple with a textual data type such as \texttt{rdfs:label} or \texttt{yago:hasPreferredMeaning}, one or more such pairs are created by tokenizing and stemming the text in the object literal. The matrix $D$ is then factorized into $D \approx AV$ with A and V being the latent-component representations of entities and attributes respectively. Despite the advantage that this approach has for handling multi-valued literals, it does not consider the sequence of words of the literal values. Note that Extended RESCAL represents RDF(S) data in such a way that there is no distinction drawn among A-Box and T-Box, i.e., both classes and instances are modeled equally as entities in a tensor. The T-Box is rather taken as soft constraints instead of letting them impose hard constraints on the model.

\noindent 
{\bf Jointly(Disp)}
is an approach which jointly learns embeddings of KGs and a text corpus of entity descriptions, i.e, it uses an alignment model to make sure the entities, relations, and words are represented in the same vector space. This approach consists of three components, namely, knowledge model, text model, and alignment model. The knowledge model is used to capture the semantics of the structured information from the KG. Given a triple $<h,r,t>$, the model defines the plausibility of the triple, same as in {\cite{wang2014knowledge}}:
\begin{equation} \label{kmodel}
Pr(h|r,t) = \frac{exp\{z(h,r,t)\}}{\sum_{\tilde{h}\in I} exp\{z(\tilde{h},r,t)\}},
\end{equation}

\noindent 
where $z(h,r,t) = b-0.5 \cdot \lVert h +r - t \rVert_2^2$,  $b=7$.  Analogously, $Pr(r|h,t)$ and $Pr(t|h,r)$ are defined. 
\\

Then, the loss function of the knowledge model is defined as follows: 
\begin{equation}
{\begin{split}
L_{K} = \sum_{(h,r,t)}[ \log Pr(h|r,t) + \log Pr(t|h,r)  \\
+ \log Pr(r|h,t)].
\end{split}}
\end{equation}

The text model adopts the same assumption made in {\cite{wang2014knowledge}} that is if two words occur in the same context then there is a relation between them. Based on this assumption, the text model defines the probability of a pair of words $w$ and $v$ co-occurring in a text window as follows:
\begin{equation}
Pr(w|v) = \frac{exp\{z(w,v)\}}{\sum_{\tilde{w}\in V} exp\{z(\tilde{w},v)\}},
\end{equation}
\noindent 
where $z(w,v) = b-0.5 \cdot \lVert \textbf{w} - \textbf{v} \rVert_2^2$. 
Then, the loss function of the text model is given as:
\begin{equation}
L_{T} = \sum_{(w,v)}\log Pr(w|v).
\end{equation}
The role of the third component, the alignment model, is to put the embeddings of the entities, relations, and words into the same vector space. This submodel works by utilizing entity descriptions to align these embeddings. For every word $w$ in the description of entity $e$,  the conditional probability of predicting $w$ given $e$ is defined as :
\begin{equation} \label{amodel}
Pr(w|e) = \frac{exp\{z(e,w)\}}{\sum_{\tilde{w}\in V} exp\{z(e,\tilde{w})\}},
\end{equation}
\noindent 
where $z(e,w) = b-0.5 \cdot \lVert \textbf{e} - \textbf{w} \rVert_2^2$. The entity vector $\textbf{e}$ in Eq~{\ref{amodel}} is the same as the entity vector appearing in Eq~{\ref{kmodel}}, i.e., an entity has a single unified representation which captures the semantics from both the structured KG and the entity descriptions. $Pr(e|w)$ is defined analogously.
Based on the definition given in Eq~{\ref{amodel}}, the loss function of the alignment model is defined as:
\begin{equation}
L_{A} = \sum_{e\in \mathcal{E}} \sum_{w\in D_{e}}[\log Pr(w|e) + \log Pr(e|w)],
\end{equation}
where $\mathcal{E}$ and $D_{e}$ denote the set of entities and the description of the entity $e$ respectively.

By adopting the joint embedding framework in {\cite{wang2014knowledge}}, the main loss of Jointly(desp) is defined as follows:
\begin{equation}
L(\{e_{i}\}, \{r_{j}\}, \{w_{l}\}) = L_K + L_T + L_A.
\end{equation}

\noindent{\bf DKRL} extends TransE~\cite{bordes2013translating} by utilizing the descriptions of entities. For each entity $e$, two kinds of vector representations are learned, i.e., structure-based $e_s$ and description-based $e_d$. These two kinds of entity representations are learned simultaneously into the same vector space but not forced to be unified so that novel entities with only descriptions can be represented. In order to achieve this, given a certain triple $<h,r,t>$ the energy function of the DKRL model is defined as: 
\begin{equation}
\begin{split}
    E = ||h_s + r - t_s|| + ||h_d + r - t_d|| \\
    + ||h_s + r - t_d|| + ||h_d + r - t_s||,
\end{split}
\end{equation}
where $h_s$ and $t_s$ are the structure-based representations, and $h_d$ and $t_d$ are the description-based representations of their corresponding entities.

In order to learn structure-based representations, the TransE approach is directly applied which considers the relation in a triple as the translation from the head entity to the tail entity. On the other hand, Continuous Bag of Words (CBOW) and a deep Convolutional Neural Network (CNN) model have been used to generate the description-based representations of the head and tail entities. In case of CBOW, short text is generated from the description based on keywords and their corresponding word embeddings are summed up to generate the entity embedding. In the CNN model, after preprocessing the description, pretrained word vectors from Wikipedia are provided as input. This CNN model has five layers and after every convolutional layer pooling is applied to decrease the parameter space of CNN and filter noises. Max-pooling is applied for the first pooling and mean pooling for the last one. The activation function used is either tanh or ReLU. The CNN model works better than CBOW because it preserves the sequence of words. 

In order to train DKRL, the following margin-based score function is considered as an objective function and minimized using a standard back propagation using stochastic gradient descent (SGD)

\begin{equation}\label{eq:dkrl}
\begin{split}
  L = \sum_{(h,r,t) \in T}\sum_{(h',r',t') \in T'} max(\gamma + d(h+r, t)\\
  - d(h'+ r', t'),0),  
\end{split}
\end{equation}
where $\gamma > 0$ is a margin hyperparameter, $d$ is a dissimilarity function and $T'$ is the set of corrupted triples. The representation of the entities can be either structure-based or description-based. 

\noindent 
{\bf Jointly} {\cite{ijcai2017xu}} learns KG embeddings by leveraging entity descriptions. Specifically, it learns a joint embedding of an entity by combining its structure-based and description-based representations with a gating mechanism. The gate is used to find balance between the structure-based and the description-based representations. For a certain entity a representation can be encoded from its descriptions by converting the description into fixed length vector. In Jointly, different text encoders have been used such as bag-of-words, LSTM, and Attentive LSTM.

For an entity $e$, its joint representation \textbf{e} is a linear interpolation between its structure-based representation $(\textbf{e}_{\textbf{s}})$ and description-based representation $(\textbf{e}_{\textbf{d}})$, which is defined as:
\begin{equation}
   \textbf{ e }= g_{e} \odot \textbf{e}_{\textbf{s}} + (1 - g_{e}) \odot \textbf{e}_{\textbf{d}},
\end{equation}
where $\odot$ is an element-wise multiplication and $g_{e}$ is a gate to balance the two information sources (structure and text) which is computed as $g_{e}= \alpha(\tilde{g_{e}})$ with $g_{e}= \tilde{g_{e}} \in \mathcal{R}^d$ being real-value vector stored in a lookup table.

The entity descriptions are encoded using either bag-of-words, LSTM, or Attentive LSTM (ALSTM) encoders in order to generate text-based representation for the corresponding entities. On the other hand, to better model the structure-based embedidngs, entities and relations can be pre-trained with any existing KG embedding models, such as TransE.

Jointly's score function is inspired by TransE and defined as follows:

\begin{equation}
   {\begin{split}
   f(h,r,t;d_{h}, d_{t})= \lVert ( \textbf{g}_{h}  \odot \textbf{h}_{s} + (1 - \textbf{g}_{h}) \\ 
   \odot  \textbf{h}_{d}) + r - ( g_{t} \odot \textbf{h}_{t} + (1 - \textbf{g}_{t}) \odot \textbf{t}_{d}) \rVert _{2}^{2}.
   \end{split}}
\end{equation}
where $\textbf{h}_{s}$, $\textbf{h}_{d}$, and $\textbf{g}_h$ are the head entity's structure-based embedding, description-based embedding, and gate respectively whereas $\textbf{t}_{s}$, $\textbf{t}_{d}$, and $\textbf{g}_{t}$ are the tail entity's structure-based embedding, description-based embedding, and gate respectively.

\noindent 
{\bf SSP} (Semantic Space Projection) {\cite{xiao2017ssp}} is a joint embedding model which learns from both structured/symbolic triples and textual descriptions. Differently from DKRL and Jointly(Desp), where first-order constraints which are weak in capturing the correlation of textual descriptions and symbolic triples are applied, SSP follows the principle that triple embedding is considered always as the main procedure and textual descriptions must interact with triples in order to learn better representation. Therefore, triple embedding is projected onto a semantic subspace such as a hyperplane to allow strong correlation by adopting quadratic constraint.

SSP applies the following scoring function: 
\begin{equation}
   f_{r} (h,t)= -\lambda \lVert \textbf{e} -\textbf{s}^{T}\textbf{es} \rVert _{2}^{2} +  \lVert \textbf{e} \rVert  _{2}^{2},
\end{equation}

where
\begin{equation}
   \textbf{e} \doteq \textbf{h} + \textbf{r} - \textbf{t},
\end{equation}

and
\begin{equation}
   \textbf{s} \doteq \frac{\textbf{s}_{\textbf{h}} + \textbf{s}_{\textbf{t}}}{\lVert \textbf{s}_{\textbf{h}} + \textbf{s}_{\textbf{t}} \rVert }. 
\end{equation}
Note that $\lambda$ is a suitable hyper-parameter, \textbf{h} and \textbf{t} are the structure (symbolic triples) based embedding of the head and tail entities respectively, $\textbf{s}_{\textbf{h}}$ and $\textbf{s}_{\textbf{t}}$ are the semantic vectors generated from the textual descriptions of the head and tail entities respectively. SSE adopts the Non-negative Matrix Factorization (NMF) topic model to generate description-based semantic vectors for entities ($\textbf{s}_{\textbf{h}}$ and $\textbf{s}_{\textbf{t}}$), i.e., by treating each entity description as a document and taking the topic distribution of the document as the representation of the corresponding entity.

SSP provides two different settings for training which are  referred to as \textbf{Std} and \textbf{Joint}. In Std, a pre-trained topic model with NMF is used to obtain description-based semantic vectors. These description-based vectors are fixed during training but the other parameters are optimized.  On the other hand, in the Joint setting the topic model is also learnt simultaneously with the KG embeddings instead of using a fixed pre-trained vectors. 

\noindent{\bf KDCoE} focuses on the creation of an alignment between entities of multilingual KGs by creating new Inter-Lingual Links (ILLs) based on an embedding approach which exploits entity descriptions. The model uses a weakly aligned multilingual KG for semi-supervised cross-lingual learning. It performs co-training of a multilingual KG embedding Model (KGEM) and a multilingual entity Description Embedding Model (DEM) iteratively in order for each model to propose a new ILL alternately. KGEM is composed of two components, i.e., a knowledge model and an alignment model, to learn embeddings based on structured information from the KGs (the non attributive triples). Given a set of languages $\mathcal{L}$, a separate k$_1$-dimensional embedding space $\mathbb{R}_L^{k_1}$ is used for each language $L \in \mathcal{L}$ to represent the corresponding relations $R_L$ and entities $E_L$.  In order to learn the embeddings for $R_L$ and $E_L$, the knowledge model adopts TransE and thus uses hinge loss as its objective function. On the other hand, a linear-transformation-based technique which has the best performance in case of cross-lingual inferences is adopted for the alignment model. This technique employs the following objective function:
\begin{equation}\label{eq:kdcoe}
S_A = \sum_{(e,e') \in I(L_i, L_j)} \lVert M_{ij}\textbf{e} -\textbf{e}'  \rVert_2,
\end{equation}
where $I(L_i, L_j)$ is ILLs between the languages $L_i$ and $L_j$, and $M_{ij}$ is a $k_1 \times k_1$ matrix used as a linear transformation on entity vectors from $L_i$ to $L_j$.

Let $S_K$ be the hinge loss function used by the knowledge model, the KGEM model then minimizes $S_{KG} = S_K + \alpha S_A$, where $\alpha$ is a positive hyperparameter. In case of DEM model, an attentive gated recurrent unit encoder (AGRU) is used to encode the multilingual entity descriptions. DEM applies multilingual word embeddings in order to capture the semantic information of multilingual entity descriptions from the word level. The two models, i.e.,  KGEM and DEM, are iteratively co-trained in order for each model to propose a new ILL alternately. 

\noindent{\bf KGloVe with literals} is an experimental attempt to incorporate entity descriptions in KGloVe KG embedding approach. The experiment is conducted on DBpedia considering the abstracts and comments of entities as their descriptions. The main goal is to extract named entities from the textual description and for every entity in the text, to replace those words representing it with the entity itself and then take its neighbouring words and entities as its context. The approach works by creating two co-occurrence matrices independently and then by merging them at the end so that a joint embedding can be performed. The first matrix is generated using the same technique as in KGloVe \cite{cochez2017global}, i.e.,  by performing Personalized PageRank (PPR) on the (weighted) graph followed by the same optimisation used in the GloVe \cite{pennington2014glove} approach.

In order to create the second matrix, the Named Entity Recognition (NER) task is performed on the entity description text using the list of entities and predicates of the KG as an input. The NER step employs a simple exact string matching technique which leads to numerous drawbacks such as missing entities due to having different keywords with the same semantics. All the English words that do not match any entity labels are added to the entity-predicate list. Then GloVe co-occurrence for text is applied to the modified text (i.e., DBpedia abstract and comments) using the entity-predicate and word list as input. Finally, the two co-occurrence matrices are summed up together to create a single unified matrix. The proposed approach has been evaluated on classification and regression tasks and the result indicates that for most of the classifiers used, except SVM, the approach does not bring significant improvement to KGloVe. However, the approach can be improved using parameter tuning with extensive experiments. 
    
\paragraph{\textbf{Summary}}
The basic differences between these models lie in the methods used to exploit the information given in the text literals and combine them with structure-based representation. One major advantage of KDCoE over text literal based embedding models is that it considers descriptions present in multilingual KGs. Also, both DKRL and KDCoE embedding models are designed to perform well for the novel entities, which have only attributive triples in the KGs. 
Jointly(Desp) aligns KG embeddings and word embeddings on word level, which may lead to losing some semantic information on phrase or sentence level. Jointly applies a gating mechanism which allows to automatically find a balance between the structural and textual information. It also uses an LSTM encoder which enables the model to select the most related information for an entity from its text description according to different relations. Unlike DKRL and Jointly(Desp), SSP focuses on characterizing the stronger correlations between entity descriptions and structured triples by projecting triple embedding onto a semantic subspace such as a hyper-plane, as discussed above. 
Another common drawbacks among the presented approaches with text literals is they focus mostly on descriptions, which is long natural language text and thus, other types of text literals, such as, names,  labels, titles, etc. are not widely considered. Moreover, another way to compare these approaches is by looking at their model complexity. Table \ref{tab:complexity_text} presents the complexity of these models in terms of their number of parameters. 

\begin{table}[]
    \centering
    \caption{
    Complexity of the models with text literals in terms of the number of parameters. $\Theta$ is the number of parameters in the base model, H is the entity embedding size, $N_d$ is the number of data relations, $L$ is the number of attribute-value pairs, $N_r$ is the number of relations, $N_w$ is the number of words, $H'$ is the word embedding size, $N_{0}^{(1)}$ is the dimension of input vectors at the first layer, $N_{1}^{(1)}$ is the dimension of output vectors at layer 1, $K$ is window size, $N_{0}^{(2)}$ is the dimension of input vectors at the second layer, $N_{1}^{(2)}$ is the dimension of output vectors at second layer, $N_{e_{1}}$ and $N_{e_{2}}$ denote the number of entities in two different languages of a multilingual KG, $N_{r_{1}}$ and $N_{r_{2}}$ denote the number of relations in two different languages of a multilingual KG, $N$ is the total number of entities and relations, and $M$ is the total number of entities, relations and words.}
    
    \begin{tabular}{c|c}
    \hline
       Model  &  \#Parameter\\
       \hline
       Extended Rescal  &  $\Theta + HL$ \\
       Jointly(Desp) & $\Theta + N_{w}H'$ \\
       DKRL & $\Theta + N_{w}H' + N_{0}^{1}KN_{1}^{(1)} + N_{0}^{(2)}KN_{1}^{(2)} $\\
       Jointly(ALSTM) & $ \Theta + (2H + 4)H$ \\
       SSP & $ \Theta + (N_{e} + N_{w})H$\\
       KDCoE &  $(N_{e_{1}} + N_{e_{2}} + N_{r_{1}} +N_{r_{2}} + H)H$  \\
       & $+ (5H' + 3N_{w})H'$ \\
       KGlove with literals & $ (N+1)N + N_w + M $\\
    \hline
    \end{tabular}
    \label{tab:complexity_text}
\end{table}

\subsection{Models with Numeric Literals}\label{numeric}
In this section, the analysis of the presented KG embedding models which use numeric literals, namely, MT-KGNN \cite{DBLP:journals/corr/abs-1708-0482}, KBLRN \cite{GarcaDurn2018KBlrnEL},  LiteralE \cite{LiteralE2019}, and TransEA \cite{Wu2018KnowledgeGE} are presented followed by a summary. Moreover, in order to show the differences between the models based on complexity, the number of parameters of each model is presented in Table \ref{table:complexity_num}. 

\paragraph{{\bf MT-KGNN}}is an approach for both relational learning and non-discrete attribute prediction on knowledge graphs in order to learn embeddings for entities, object properties, and data properties. It is composed of two networks, namely, the Relational Network (RelNet) and the Attribute Network (AttrNet). RelNet is a binary (pointwise) classifier for 
triple 
prediction whereas AttrNet is a regression task for attribute value prediction. Given $n$, $m$, and $l$ as entity, relation, and literal embedding dimensions respectively, the model passes as an input $[e_{i} , r_{k}, e_{j}, t]$ to RelNet and $[a_{i} ,v_{i} , a_{j} ,v_{j}]$ to AttrNet, where $e_{i}$ , $e_{j} \in R^n$, $r_{k} \in R^{m}$,  $t$ is the classification target which is 0 or 1, $a_{i}, a_{j} \in R^l$, and $v_{i}$ and $v_{j}$ are normalized continuous values in the interval $[0, 1]$. Note that the inputs to AttrNet, i.e.,  $[a_{i} ,v_{i} , a_{j} ,v_{j}]$, are taken from attributive triples with non-discrete literal values. An embedding lookup layer is used to retrieve the corresponding vector representations given these inputs as one-hot encoded indices. 

In RelNet, a concatenated 
triple 
is passed through a nonlinear transform and then a sigmoid function is applied to get a linear transform:
\begin{equation}
\begin{split}
    g_{rel}(e_{i} , r_{k}, e_{j}) = \sigma(\overrightarrow{w}^T f(W_d^T[\overrightarrow{e_{i}};\overrightarrow{e_{j}};\overrightarrow{r_{k}}]) \\
    + b_{rel}),
\end{split}
\end{equation}
where $w \in R^{h\times1}$ and $W_d \in R^{3n\times h}$ are parameters of the network. $\sigma$,  $f$, and $b_{rel}$ are the sigmoid function, the hyperbolic tangent function tanh,  and a scalar bias respectively. RelNet is trained by minimizing the following cross entropy loss function:
\begin{equation}
\begin{split}
    L_{rel} = -\sum_{i=1}^{N}t_i \log g_{rel}(\xi_i)\\
    +(1-t_i)\log(1-g_{rel}(\xi_i))),
\end{split}
\end{equation}
where $\xi_i$ denotes triple $i$ in batch of size $N$ and $t_i$ takes the value $0$ or $1$.
In case of AttrNet, two regression tasks are performed, one for the head data properties and another for those of the tail. The following scoring functions are defined for these two tasks:
\begin{equation}
    g_h(a_i) =  \sigma(\overrightarrow{u}^T f(B^T[\overrightarrow{a_i};\overrightarrow{e_i}])
    +b_{z_1}), 
\end{equation}
and
\begin{equation}
    g_t(a_j) =  \sigma(\overrightarrow{y}^T f(C^T[\overrightarrow{a_j};\overrightarrow{e_j}])+b_{z_2}),
\end{equation}
where $u, y \in R^{h_a\times 1}$ and $B, C \in R^{2n\times h_a}$  are parameters of AttrNet. $h_a$ is the size of the hidden layer and $b_{z_1}$, $b_{z_2}$ are scalar biases.
Each AttrNet is trained by optimizing Mean Squared Error (MSE) loss function: 
\begin{equation}
    MSE(s,s^*) = \frac{1}{N}\sum_{i=1}^{N}(s_i - s_i^*)^2.
\end{equation}
where $s$ and $s^*$ are predicted labels (scores computed by the model ) and ground truth labels respectively. 
\\
The overall loss of the AttrNet is computed by adding the MSE of the head AttrNet and that of the tail AttrNet as follows: 
\begin{equation}
\begin{split}
    L_{attr} = MSE(g_h(a_i),(a_i)^*) \\
    + MSE(g_t(a_j),(a_j)^*),
\end{split}
\end{equation}
where $(a_i)^*, (a_j)^*$ are the ground truth labels. Finally, the two networks are trained in a multi-task fashion using a shared embedding space.

\noindent{\bf KBLRN} works by combining relational (R), latent (L), and numerical (N) features together. The model is designed mainly for the purpose of KG completion. It uses a probabilistic PoE (Product of Experts) method to combine these feature types and train them jointly end to end. Each relational feature is formulated as a logical formula, by adopting the rule mining approach AMIE+ \cite{galarraga2015fast}, to be evaluated in the KB to compute the feature's value. The latent features are the ones that are usually generated using an embedding approach such as DistMult. Numerical features are used with the assumption that, for some relation types, the differences between the head and tail can be seen as characteristics for the relation itself. Given a triple $d = (h, r, t)$, for each (relation type $r$, and feature type $F \in \{L, R, N\}$) pair, individual experts are defined based on linear models and DistMult embedding method as follows: 
 \begin{equation}
    f_{(r,L)}(d \ | \ \theta_{(r,L)}) = exp((e_h * e_t) \cdot w^r),
\end{equation}
\begin{equation}
    f_{(r,R)}(d \ | \ \theta_{(r,R)}) = exp(r_{(h,t)} \cdot w_{rel}^r),
\end{equation}
\begin{equation}
    f_{(r,N)}(d \ | \ \theta_{(r,N)}) = exp(\phi(n_{(h,t)}) \cdot w_{num}^r), 
\end{equation}
and
\begin{equation}
    f_{(r^\prime,F)}(d \ | \ \theta_{(r^\prime,F)}) = 1 \ for \ all \ r^\prime \neq  r 
\end{equation}
\noindent
where $w_{r},w_{rel}^r ,w_{num}^r$ are the parameter vectors for the latent, relational, and numerical features corresponding to the relation r. Also, * is the element-wise product, $\cdot$ is the dot product, and $\phi$ is the radial basis function (RBF) applied element-wise to the differences of values $n_{(h,t)}$ computed as follows:
\begin{equation}
\begin{split}
    \phi(n_{(h,t)}) = [exp(\frac{-\lVert n_{(h,t)}^{(1)} - c_1\rVert _2^2
    }{\sigma_1^2}) \dots \\ exp(\frac{-\lVert n _{(h,t)}^{(d_n)} - c_{d_n}\rVert_2^2}{\sigma_{d_n}^2})].
\end{split}
\end{equation}
 Here, $d_n$ corresponds to the relevant numerical features.
A PoE's probability distribution for a triple $d=(h, r, t)$ is defined as follows:
\begin{equation}
    p(d \ | \ \theta_1 \dots \theta_n) = \frac{\Pi_F f_{(r,F)}(d \ | \ \theta_{(r,F)})}{\sum_c \Pi_F f_{(r,F)}(c \ | \ \theta_{(r,F)})},
\end{equation}
where c denotes all possible triples.
The parameters of the entity embedding model are shared by all the experts in order to create dependencies between them. In this approach, the PoE are trained with negative sampling and a cross entropy loss to give high probability to observed triples. 

\noindent{\bf LiteralE} incorporates literals into existing latent feature models designed for link prediction.  In this approach, without loss of generality, the focus lies on incorporating numerical literals into three state-of-the-art embedding methods: DistMult, ComplEx, and ConvE. Given a base model, for instance Distmult, LiteralE modifies the scoring function $f$ used in Distmult by replacing the vector representations of the entities $e_{i}$ in $f$ with literal enriched representations $e_{i}^{lit}$. In order to generate $e_{i}^{lit}$, LiteralE uses a learnable transformation function $g$ which takes $e_{i}$ and its corresponding literal vectors $l_i$ as inputs and maps them to a new vector. The function $g$ is defined, as shown below, based on the concept of GRU in order to make it flexible, learnable, and capable to decide, if it is beneficial to incorporate the literal information or not:

\begin{equation}
g : \mathbb{R}^H \times \mathbb{R}^{N_d} \rightarrow \mathbb{R}^H,
\end{equation}
and
\begin{equation}
\textbf{e},\textbf{l} \mapsto \textbf{z} \odot \textbf{h} + (1 - \textbf{z}) \odot \textbf{e},
\end{equation}
where   
\begin{equation}
\textbf{z} : \sigma(\textbf{W}_{ze}^T\textbf{e} + \textbf{W}_{zl}^T\textbf{l} + \textbf{b}),
\end{equation}
and
\begin{equation}
\textbf{h} = h(\textbf{W}_h^T[\textbf{e},\textbf{l}]).
\end{equation}
Note that $\textbf{W}_{ze} \in \mathbb{R}^{H \times H} ,  \textbf{W}_{zl} \in \mathbb{R}^{N_d \times H}, \textbf{b} \in \mathbb{R}^H$, and $\textbf{W}_h \in \mathbb{R}^{H+N_d \times H} $ are the parameters of $g$, $\sigma$ is the sigmoid function, $\odot$ denotes the element-wise multiplication, and h is a component-wise nonlinearity. The scoring function $f(\textbf{e}_i, \textbf{e}_j, \textbf{r}_k)$ has been replaced with $f(g(\textbf{e}_i , \textbf{l}_i ), g(\textbf{e}_j , \textbf{l}_j ), \textbf{r}_k)$ and trained following the same procedure as in the base model.

\noindent{\bf TransEA} has two component models; a directly adopted translation-based structure embedding model (i.e., TransE) and a newly proposed attribute embedding model. In the former, the scoring function of a given triple $<h,r,t>$, is defined as follows:
\begin{equation}
f_r(h,t) = -\lVert h + r -t\rVert _{1/2},
\end{equation}
where $||x||_{1/2}$ denotes either the L1 or L2 norm. The loss function of the structure embedding, for all the relational 
triples 
in the KG, is defined as:
\begin{equation}
{\begin{split}
L_R = \sum_{<h,r,t> \in T}\sum_{<h',r,t'> \in T'} max(\gamma + f_r(h,t) \\
-f_r(h',t'), 0) , 
\end{split}}
\end{equation}
where $T'$ denotes the set of negative 
triples 
constructed by corrupting either the head or the tail entity and $\gamma > 0$ is a margin hyperparameter.

For the attribute embedding, it uses all attributive triples containing numeric values as input and applies a linear regression model to learn embeddings of entities and attributes. Given an attributive triple $<e, a, v>$, the scoring function is defined as: 
\begin{equation}
f_a(e,v) = -\lVert \textbf{a}^T\cdot \textbf{e} + b_a -v\rVert _{1/2},
\end{equation}
where $\textbf{a}$ and $\textbf{e}$ are vectors of attribute $a$ and entity $e$, $b_a$ is a bias for attribute $a$. On the other hand, given all the attributive triples with numeric values $S$, the loss function for the attributive embedding is computed as:
\begin{equation}
L_A = \sum_{<e,a,v> \in S} f_a(e,v), 
\end{equation}

The main loss function for TransEA ($i.e., L = (1 -\alpha) \cdot L_R + \alpha \cdot L_A$)  is defined by taking the sum of the respective loss functions of the component models with a hyperparameter to assign a weight for each of the models. Finally, the two models are jointly optimized in the training process by sharing the embeddings of entities.

\paragraph{\textbf{Summary}} 
Despite their support for numerical literals, all the embedding methods discussed fail to interpret the semantics behind units/data typed literals. For instance, given the following two triples taken from DBpedia,
\begin{lstlisting}[basicstyle=\footnotesize\ttfamily] 
<http://dbpedia.org/resource/Anton_Baraniak, dbp:weight, "110.0"^^<http://dbpedia.org/datatype/kilogram>,
<http://dbpedia.org/resource/Katelin_Snyder, dbp:weight, "110.0"^^<http://dbpedia.org/datatype/pound>
\end{lstlisting}
the literal value "110.0" from the first triple and the literal value "110.0" from the second triple could be considered exactly the same if the semantics of the types kilogram and pound are ignored. Moreover, most of the models do not have a proper mechanism to handle multi-valued literals. 

Regarding model complexity, the number of parameters used in each model is presented in Table \ref{table:complexity_num} to show the complexity in terms of the parameters. It is noted that the complexity of the models depend on the size of the dataset and TransEA has lower complexity as compared to the other models. 

\begin{table}[ht]
    \centering
    \caption{Complexity of the models with numerical literals in terms of the number of parameters. $\Theta$ is the number of parameters in the base model, H is the entity embedding size, $N_d$ is the number of data relations, $\Lambda$ is the size of the hidden layer in the Attrnet networks of MTKGNN, $N_r$ is the number of relations, and M is attribute embedding size.}
    \begin{tabular}{cc}
    \hline
        Model &  \#Parameters\\
        \hline
        LiteralE with g &  $\Theta$ + $2H^2$ + 2$N_d$H + H \\
        LiteralE with $g_{lin}$ & $\Theta$ + ($N_d$H + H)H\\
        MTKGNN &  $\Theta$ + $N_d$H + 2(2H$\Lambda$ + $\Lambda$) \\
        KBLN &  $\Theta$ + $N_rN_d$ \\ 
        TransEA &  $\Theta$ + $N_{d}M$ \\
    \hline
    \end{tabular}
    \label{table:complexity_num}
\end{table}

\subsection{Models with Image Literals}\label{image}
\vspace{0.25cm}
In this section, KG embedding models utilizing images of entities, namely, IKRL {\cite{Xie2017ImageembodiedKR}} and MTKGRL {\cite{mousselly2018multimodal}} are discussed. First, a detailed analysis of the models is presented followed by a summary. Moreover, in order to show the differences between the models based on complexity, the number of parameters of each model is presented in Table {\ref{tab:complexity_image}}. 

\paragraph{{\bf IKRL}}\cite{Xie2017ImageembodiedKR} learns embeddings for KGs by jointly training a structure-based representation with an image-based representation. The structure-based representation of an entity is learned by adapting a conventional embedding model like TransE. For the image-based representation, given the fact that an entity may have multiple image instances, an image encoder is applied to generate an embedding for each instance of a multi-valued image relation. The image encoder consists of a neural representation module and a projection module to extract discriminative features from images and to project these representations from image space to entity space respectively. 

For the i-th image, its image-based representation $p_i$ in entity space is computed as:
\begin{equation}
\textbf{p}_i = \textbf{M} \cdot f(img_i),
\end{equation}
where $M \in \mathbb{R}^{d_i \times d_s}$ is the projection matrix with $d_i$ and $d_s$ representing the dimension of image features and the dimension of entities respectively. $f(img_i)$ is the i-th image feature representation in image space.

Attention-based multi-instance learning is used to integrate the representations learned for each image instance by automatically calculating the attention that should be given to each instance. The attention for the i-th image representation $p_i^{(k)}$ of the k-th entity is given as:
\begin{equation}
att(\textbf{p}_i^{(k)}, \textbf{e}_S^{(k)}) = \frac{exp(\textbf{p}_i^{(k)} \cdot \textbf{e}_S^{(k)})}{\sum_{j=1}^n exp(\textbf{p}_j^{(k)} \cdot \textbf{e}_S^{(k)})},
\end{equation}
where $\textbf{e}_S^{(k)}$ denotes the structure-based representation of the k-th entity. The higher the attention the more similar the image-based representation is to its corresponding structure-based representation which indicates that it should be given more importance when aggregating the image-based representations.  The aggregated image-based representation for the k-th entity is defined as follows:
\begin{equation}
\textbf{e}_I^{(k)} = \sum_{i=1}^n\frac{att(\textbf{p}_i^{(k)}, \textbf{e}_S^{(k)}) \cdot \textbf{p}_i^{(k)}}{\sum_{j=1}^n att(\textbf{p}_j^{(k)}, \textbf{e}_S^{(k)})}.
\end{equation}

Given a triple, the overall energy function is defined by combining four energy functions (i.e., $E(h,r,t) = E_{SS} + E_{II} + E_{SI} + E_{IS}$. These energy functions are based on two kinds of entity representations (i.e, structure-based and image-based representations). The first energy function (i.e., $E_{SS} = \lVert h_S + r - t_S\rVert$) is same as TransE and the second function (i.e., $E_{II} = \lVert h_I + r - t_I \rVert$) uses their corresponding image-based representations for both head and tail entities. The third function (i.e., $E_{SI} = \lVert h_S + r - t_I \rVert$) is based on the structure-based representation of the head entity and the image-based representation of the tail entity whereas the fourth function (i.e., $E_{IS} = \lVert h_I + r - t_S \rVert$) is the exact opposite. These third and forth functions ensure that both structure-based representation and image-based representations are learned into the same vector space.

Given the energy function $E(h,r,t)$, a margin-based scoring function is defined as follows: 
\begin{equation}
\begin{split}
   L = \sum_{(h,r,t)\in T} \sum_{(h',r',t') \in T'} max(\gamma + E(h,r,t)\\
-  E(h',r',t'), 0),
\end{split}
\end{equation}
where $\gamma$ is a margin hyperparameter and $T'$  is the negative sample set of T generated by replacing the head entity, tail entity or the relation for each triple in T. Note that triples which are already in T are removed from $T'$. 

\noindent 
\textbf{MTKGRL} {\cite{mousselly2018multimodal}} is a KG embedding approach which combines structural (symbolic), visual, and linguistic KG representations. The structural representations are created by adopting TransE embedding technique whereas visual embeddings are obtained from the feature layers of deep networks for image classification on the images that are associated with entities. For linguistic representations, pre-trained word embedding technique, specifically the skipgram model, is used. However, the information source for the linguistic representation are not literals from the KG but an external source, i.e., the word embedding model, trained on Google 100B token news dataset. Due to this fact, the model MTKGRL is not considered as multi-modal KG embedding model in the context of this survey and thus, it is not categorized under `Models with Multi-modal Literals' (Sec~{\ref{multi-modal}}).

MTKGRL defines an energy function for each kind of representation and also their combinations, i.e., structural energy, multimodal energies, and structural-multimodal energies. Structural energy is adopted from TransE, which is defined as \textit{$ E_{S} = \lVert \textbf{h}_{\textbf{s}} + \textbf{r}_{\textbf{s}} - \textbf{t}_{\textbf{s}}  \rVert$}. The multimodal representations for the head and tail entities are computed as \textit{${\textbf{h}_{\textbf{m}} = \textbf{h}_{\textbf{w}} \oplus \textbf{h}_{\textbf{i}}}$} and \textit{${\textbf{t}_{\textbf{m}} =  \textbf{t}_{\textbf{w}} \oplus \textbf{t}_{\textbf{i}} }$} respectively, where the operator $\oplus$ can be a concatenation operator or a mapping function.

The multimodal energy function under the translational assumption is given as:
\begin{equation}
{E_{M1} = \lVert \textbf{h}_{\textbf{m}} + \textbf{r}_{\textbf{s}} - \textbf{t}_{\textbf{m}}  \rVert }. 
\end{equation}
$E_{M1}$ can be extended by considering the structural embeddings in addition to the multimodal embeddings as follows:
\begin{equation}
{E_{M2} = \lVert (\textbf{h}_{\textbf{m}} + \textbf{h}_{\textbf{s}}) \textbf{r}_{\textbf{s}} - (\textbf{t}_{\textbf{m}} + \textbf{t}_{\textbf{s}})  \rVert }. 
\end{equation}
On the other hand, in order to allow the structural and multimodal embeddings to be learned in the same vector space, the following structural-multimodal energies are defined as shown below:
\begin{subequations}
\begin{align}
{E_{SM} = \lVert \textbf{h}_{\textbf{s}} + \textbf{r}_{\textbf{s}} - \textbf{t}_{\textbf{m}}  \rVert } 
\\
{E_{MS} = \lVert \textbf{h}_{\textbf{m}} + \textbf{r}_{\textbf{s}} - \textbf{t}_{\textbf{s}}  \rVert } 
\end{align}
\end{subequations}

The overall energy function, shown in Equation~{\ref{energy}}, is defined by combining the aforementioned energy functions, i.e., $E_{S}$, $E_{M1}$, $E_{M2}$, $E_{SM}$, $E_{MS}$.
\begin{equation} \label{energy}
\begin{split}
    E(h,r,t) = E_{S} + E_{M1} + E_{M2} +  E_{SM}\\
    + E_{MS}
\end{split}
\end{equation}
Finally, a margin-based ranking loss function is minimized in order to train the model.
\paragraph{\textbf{Summary}} \label{para:IKRL-Summary}
IKRL makes use of the images of entities for KG representation learning by combining structure-based representation with image-based representation. However, given a triple $<h,r,t>$, in order to achieve very good representations for the entities $h$ and $t$, both entities are required to have images associated with them. The other issue with this model is that an image is considered as an attribute of only those entities it is associated with. For example, if there is an image of two entities $e_1$ and $e_2$ but the image is associated with only $e_1$, then it will be taken as one image instance of $e_1$ but not of $e_2$. However, it would be more beneficial to explicitly associate images with all the entities they represent before using them for learning KG embedding.  
Some of the main points which make MTKGRL differ from IKRL are: i) in addition to images, MTKGRL uses linguistic embeddings for entities, ii) MTKGRL introduces an additional energy function that considers both linguistic and visual representations of entities as discussed above. These differences allow MTKGRL to learn better representation for KGs as compared to IKRL. 

\begin{table}[]
    \centering
    \caption{
    Complexity of the models with text literals in terms of the number of parameters. $\Theta$ is the number of parameters in the base model, $H$ is the entity embedding size, $H_{i}$ represents the dimension of image features, $\theta_{AlexNet}$ is the number of parameters in AlexNet {\cite{krizhevsky2012imagenet}}, $N_e$ represents the number of entities, and $N_i$ is the number of images.}
    \begin{tabular}{c|c}
    \hline
       Model  &  \#Parameter\\
       \hline
       IKRL  &  $ \Theta + H_{i}H + \Theta_{AlexNet}$ \\
       MTKGRL & $ \Theta + N_{e}H + N_{i}H_{i}$ \\
    \hline
    \end{tabular}
    \label{tab:complexity_image}
\end{table}
\subsection{Models with Multi-modal Literals}\label{multi-modal}
This section presents an analysis of the embedding models making use of at least two types of literals providing complementary information. First, the category with numeric and text literals is discussed followed by the category with numeric, text, and image. 
Moreover, in order to show the differences between the models based on complexity, the number of parameters of each model is presented in Table {\ref{tab:complexity_multimodal}}.

\subsubsection {Models with Numeric and Text Literals} 
\vspace{0.25cm}
\paragraph{{\bf LiteralE with blocking}}\cite{de2018towards} proposes to improve the effectiveness of the data linking task by combining LiteralE with a CER blocking \cite{7474218} strategy. Unlike LiteralE, given an attributive triple $<h,d,v>$, in addition to the object literal value $v$ it also takes literals from URI infixes of the head entity $h$ and the data relation $d$. The CER blocking is based on a two-pass indexing scheme. In the first pass, Levenshtein distance metric is used to process literal objects and URI infixes whereas in the second pass semantic similarity computation with WordNet~\cite{miller1995wordnet} is applied to process object/data relations. All the extracted literals are tokenized into word lists so as to create inverted indices. 
The same training procedure as in LiteralE is used to train this model. For every given triple $<h,r,t>$,  the scoring function $f$ from LiteralE is adopted to compute scores for all the triples $<h,r,t'>$ in the knowledge graph. A sigmoid function, $ p = \sigma(f(.))$ , is used to produce probabilities. Then, the model is trained by minimizing the binary cross-entropy loss of the produced probability function vector with respect to the vector of truth values for the triples.

\noindent{\bf EAKGAE} \cite{Trsedya2019entity} is an approach designed for entity alignment between KGs by learning a unified embedding space for the KGs. The entity alignment task has three main modules: Predicate alignment, Embedding learning, and Entity alignment. The predicate alignment module merges two KGs together by renaming similar predicates so as to create unified vector space for the relationship embeddings. The embedding learning module jointly learns entity embeddings of two KGs using structure embedding (by adapting TransE) and attribute character embedding. The adapted TransE is customized in a way that more focus can be given to triples with aligned predicates.
This is obtained by adding a weight $\alpha$ to control the embedding learning over the triples. Thus, the following objective function $J_{SE}$ is defined for the structure-based embedding:
\begin{equation}
J_{SE} = \sum_{t_r \in T_r} \sum_{t_r' \in T_r'}max(0, \gamma + \alpha(f(t_r)-f(t_r'))),
\end{equation}
and
\begin{equation}
\alpha = \frac{count(r)}{|T|},
\end{equation}
where $T_r$ and $T_r'$ are the sets of valid triples and corrupted triples respectively, $count(r)$ is the number of occurrences of the relation $r$, and $|T|$ is the total number of triples in the merged KG.

On the other hand, the attributing character embedding is designed to learn embeddings for entities from the strings occurring in the attributes associated with the entities. The purpose is to enable the entity embeddings from two KGs to fall into the same vector space despite the fact that the attributes come from different KGs.  The attribute character embedding is inspired by the concept of translation in TransE. 
Given a triple $(h,r,a)$, the data property r is interpreted as a translation from the head entity $h$ to the literal value $a$ i.e. $h + r = f_{a}(a)$ where $f_{a}(a)$ is a compositional function. This function encodes the attribute values into a single vector mapping similar attribute values into similar representation. Three different compositional functions SUM, LSTM, and N-gram-based functions have been proposed. SUM is defined as a summation of all character embeddings of the attribute value. In LSTM, the final hidden state is taken as a vector representation of the attribute value. The N-gram-based function, which shows better performance than the others according to their experiments, uses the summation of n-gram combination of the attribute value. 

The following objective function is defined for the attribute character embedding:
\begin{equation}
J_{CE}= \sum_{t_a \in T_a}\sum_{t_a' \in T_a'}max(0, [\gamma + \alpha(f(t_a) - f(t_a'))]),
\end{equation}
$$
T_a = {<h,r,a> h \in G}; f(t_a) = \lVert h+r-f_a(a) \rVert, $$ and 
$$ T'_a = \{<h',r,a> h' \in G\} \cup \{<h,r,a'> a' \in A\},$$
where, $T_a$ and $T_a'$ are the sets of valid attribute triples and corrupted attribute triples with A being the set of attributes in a given KG $G$. The corrupted triples are created by replacing the head entity with a random entity or the attribute with a random attribute value. Here, $f(t_a)$ is the plausibility score computed based on the embedding of the head entity h, the embedding of the relation r, and the vector representation of the attribute value obtained using one of the compositional functions $f_a(a)$.

The attribute character embedding $h_{ce}$ is used to shift the structure embedding $h_{se}$ into the same vector space by minimizing the following objective function:
\begin{equation}
J_{SIM} = \sum_{h \in G_1 \cup G_2} [1- \lVert h_{se}\rVert _2 \cdot \lVert h_{ce}\rVert _2],
\end{equation}
where, $\lVert x \rVert _2$ is the L$_{2}$-Norm of vector x. This way the similarity of entities from two KGs is captured by the structure embedding based on the entity relationships and by the attribute embedding based on the attribute values. 

All the three functions are summed up to an overall objective function J (i.e., $ J= J_{SE} + J_{CE} + J_{SIM}$) for jointly learning both structure and attribute embeddings. Finally, the alignment is done by defining a similarity equation with a specified threshold. Moreover, a transitivity rule has been applied to enrich triples in the KGs to get a better attribute embedding result.

\paragraph{\textbf{Summary}}
The common drawback with both methods (LiteralE with blocking and EAKGE) is that text and numeric literals are treated in the same way. They also do not consider literal data type semantics or multi-valued literals in their approach. Furthermore, since EAKGAE is using character-based attribute embedding, it fails to capture the semantics behind the co-occurrence of syllables.

\subsubsection{Models with Numeric, Text, and Image Literals} \vspace{0.25cm}
\paragraph{{\bf MKBE}} \cite{mmkb:emnlp18}  is a multi-modal KG embedding, in which the text, numeric and image literals are modelled together.  The main objective of this approach is to utilize all the observed subjects, objects, and relations (object properties and data properties) in order to predict whether any fact holds. It extends DistMult, which creates embedding for entities and relations, by adding neural encoders for different data types.
Given a triple $<s, r, o>$, the head entity $s$ and the relation $r$ are encoded as independent embedding vectors using one-hot encoding through a dense layer. Similarly, if the object $o$ is a categorical value, then it will be represented through a dense layer with a relu activation which has the same number of nodes as the embedding space dimension. On the other hand, if the object $o$ is rather a numerical value, then a feed forward layer, after standardizing the input, is used in order to learn embeddings for $o$ by projecting it to a higher-dimensional space. If $o$ is a short text (such as names and titles), it is encoded using character-based stacked, bidirectional GRUs and the final output of the top layer will be taken as the representation of $o$.  On the contrary, if $o$ is a long text such as entity descriptions, CNN over word embeddings will be used to get the embeddings for $o$. The object $o$ can also be an image, and in such a case, the last hidden layer of VGG pretrained network on ImageNet \cite{Simonyan14verydeep}, followed by compact bilinear pooling, is used to obtain the embedding of $o$. Given the vector representations of the entities, relations and attributes, the same scoring function from DistMult is used to determine the correctness probability of triples. 

The binary cross-entropy loss, as defined below, is used to train the model:
\begin{equation}
\sum_{(s,r)}\sum_{o}t_o^{s,r}\log(p_o^{s,r}) + (1-t_o^{s,r})\log(1-p_o^{s,r}),
\end{equation}
where for a given subject relation pair $(s, r)$, binary label vector $t^{s,r}$  over all entities is used to indicate whether $<s, r, o>$ is observed during training. $p_o^{s,r}$  denotes the model's probability  of truth for any triple $<s, r, o>$ computed  using a sigmoid function. 

Moreover, using these learned embeddings and different neural decoders, a novel multimodal imputation model is introduced to generate missing multimodal values, such as numerical data, categorical data, text, and images, from information in the knowledge base. In order to predict the missing numerical and categorical data such as dates, gender, and occupation, a simple feed-forward network on the entity embedding is used. For text, the adversarially regularized autoencoder (ARAE) has been used to train generators that decodes text from continuous codes, having the generator conditioned on the entity embeddings instead of random noise vector. Similarly, the combination of BE-GAN structure with pix2pix-GAN model is used to generate images, conditioning the generator on the entity embeddings.

\paragraph{\textbf{Summary}} Despite the attempt made in incorporating text literals, numeric literals, and images into the KG embedding, the model (MKBE) fails to capture the semantics of the data types/units of (numeric) literal values. Besides, similar to IKRL, it takes an image $I$ as an instance of a certain entity $e$ only if, $I$ is initially associated with $e$ in the  dataset considered (refer to Section~\ref{para:IKRL-Summary} for more details). 

\begin{table}[]
    \centering
    \caption{
    Complexity of the models with multimodal literals in terms of the number of parameters. $\Theta$ is the number of parameters in the base model, $H$ is the entity embedding size, $N_d$ is the number of data relations, $N_{char}$ is the number of characters, and $N_i$ is the number of images, $\Theta_{CNN}$ is the number of parameters in the CNN model used in {\cite{francis2016capturing}}, $\Theta_{ARAE}$ is the number of parameters in ARAE {\cite{zhao2017adversarially}} where instead of using the random noise vector $z$, the generator is conditioned on the entity embeddings, $\Theta_{GAN}$ denotes the sum of the number of parameters in BE-GAN {\cite{berthelot2017began}} and in pix2pix-GAN {\cite{isola2017image}}. }
    \begin{tabular}{c|c}
    \hline
       Model  &  \#Parameter\\
       \hline
       LiteralE with blocking &  $\Theta + (N_dH + H)H$\\
       EAKGAE &  $\Theta + (N_{d} + N_{char})H$\\
       MKBE & $\Theta + (2(N_d + 3(N_{char} + H)) +  N_i)H$ \\
       & $+ \Theta_{CNN} + \Theta_{ARAE} + \Theta_{GAN}  $\\
    \hline
    \end{tabular}
    \label{tab:complexity_multimodal}
\end{table}
 
\section{Applications}\label{sec:application}
This section discusses different applications of KG embeddings on which the previously described methods have been trained and/or evaluated.  

\paragraph{\textbf{Link prediction.}} In general terms, link prediction can be defined as a task of identifying missing information in complex networks \cite{linkPrediction, LinkPredSurvey}. Specifically in the case of KGs,  link prediction models aim at predicting new relations between entities leveraging the existing links for training. Along with predicting relations between the entities link prediction also focuses on the task of predicting either the head or the tail entity with respect to a relation. Then it decides if a new triple, which is not observed in the KG, is valid or not.  Formally, let $G$ be a KG with a set of entities $E = \{e_{1}, \dots, e_{n}\}$ and a set of object relations $R = \{r_{1}, \dots , r_{m}\}$, then link prediction can be defined by a mapping function $\psi : E \times E \times R \rightarrow R$ which assigns a score to every possible triple $(e_{i}, e{_j}, r_{k}) \in E \times E \times R$. A high score indicates that the triple is most likely to be true.

Link prediction is one of the most common tasks used for evaluating the performance of KG embeddings. Head prediction, tail prediction, and relation prediction are different kinds of sub-tasks related to link prediction. Head prediction aims at identifying a missing head entity where the relation and tail entity are given, and analogously for tail prediction and relation prediction. Most of the models  discussed in Section~\ref{sec:techniques} have been evaluated on some or all of these prediction tasks. Head and tail prediction are used to evaluate the models LiteralE~\cite{LiteralE2019}, TransEA~\cite{Wu2018KnowledgeGE}, KBLRN~\cite{GarcaDurn2018KBlrnEL}, KDCoE~\cite{chen2018co}, EAKGAE~\cite{Trsedya2019entity},  IKRL~\cite{Xie2017ImageembodiedKR}, MKBE~\cite{mmkb:emnlp18}, 
MTKGRL {\cite{mousselly2018multimodal}}, Jointly(Desp) {\cite{zhong2015aligning}}, Jointly {\cite{ijcai2017xu}}, and SSP {\cite{xiao2017ssp}}. 
On the other hand, DKRL~\cite{xie2016representation} has been evaluated on all kinds of link prediction tasks: head, tail, and relation predictions. 
In Extended RESCAL~\cite{nickel2012factorizing}, two kinds of link prediction experiments have been conducted on the Yago 2~\cite{Yago2}, i.e., i) tail prediction by fixing the relation type to {\tt rdf:type}, and ii) general link prediction experiments for all relation types. Unfortunately, it is not possible to compare the obtained evaluation results of all these models because the experiments have been carried out on different datasets and also different link prediction procedures have been followed. Taking this into consideration, in this survey, experiments have been conducted on head and tail prediction tasks for these models (see Section~\ref{sec:experimentation}).

\paragraph{\textbf{Triple Classification.}}
The goal of the triple classification task is the same as that of link prediction. A potential triple $<h, r, t>$ is classified as 0 (false) or 1 (true), i.e., a binary classification task. The embedding models MTKGNN \cite{xie2016representation}, IKRL \cite{Xie2017ImageembodiedKR}, 
MTKGRL {\cite{mousselly2018multimodal}}, Jointly(Desp) {\cite{zhong2015aligning}}, and Jointly {\cite{ijcai2017xu}} 
have been evaluated on this task. However, since they do not use a common evaluation dataset, it is not possible to compare the reported results directly. 

\paragraph{\textbf{Entity Classification.}}
Given a KG $G$, with a set of entities $E$ and types $T$ and with an entity $e \in E$ and type $t \in T$, the task of entity classification is to determine if a potential entity type pair $(e,t)$ which is not observed in G ($(e,t) \notin G$) is a missing fact or not. This task is an entity type  prediction using a multi-label classification algorithm considering the entity types in G as given classes. In DKRL \cite{xie2016representation}, 
Extended RESCAL~{\cite{nickel2012factorizing}}, and SSP {\cite{xiao2017ssp}}, Entity classification has been used for model evaluation. 

\paragraph{\textbf{Entity Alignment.}}
Given two KGs $G_1$ and $G_2$, the goal of the entity alignment task is to identify those entity pairs $(e_1,e_2)$ where $e_1$ is an entity in $G_1$ and $e_2$ is an entity in $G_2$ which denote the same real world entities, and hence the integration of $G_1$ and $G_2$ can be possible through these unified entities, i.e., entity pairs. Different embedding-based models have been proposed recently for the entity alignment task. Among the models that are included in this survey, EAKGAE \cite{Trsedya2019entity} and KDCoE \cite{chen2018co} have been proposed for the entity alignment task. Specifically, KDCoE \cite{chen2018co} uses a cross-lingual entity alignment task which determines similar entities in different languages. Despite the fact that both these models use the same task for evaluation, the entity alignment task, their experimental results cannot be compared since they are based on different datasets.

\paragraph{\textbf{Other Applications.}} Attribute-value prediction, near\-est-neighbor analysis, data linking, document classification, and 
relational fact extraction 
are other application scenarios used for the evaluation of the models under discussion. 
Attribute-value prediction is the process of predicting the values of (discrete or non-discrete) attributes in a KG. For example, a missing value of a person's weight can be identified using the attribute value prediction task which is commonly seen as a KG completion task. 
In MTKGNN \cite{DBLP:journals/corr/abs-1708-0482}, attribute-value prediction is applied using an attribute-specific Linear Regression classifier for evaluation. The same task has been employed in MKBE \cite{mmkb:emnlp18} for model evaluation by imputing different multi-modal attribute values.

Nearest Neighbor Analysis is a task of detecting the nearest neighbors of some given entities in the latent space learned by an embedding model. This task has been performed in LiteralE \cite{LiteralE2019} to compare DistMult+LiteralE with the base model DistMult. On the other hand, data linking and document classification tasks have been used in LiteralE with blocking \cite{de2018towards} and KGlove with literals \cite{cochez2018first} respectively (refer to \cite{de2018towards} and \cite{cochez2018first} for more details). 
Relational fact extraction is a task of extracting facts/triples from plain text and has been used as a model evaluation task in Jointly(Desp) {\cite{zhong2015aligning}}. 
Table~\ref{table:application-summary} summarizes all the applications on which the KG embedding models with literals have been evaluated.

\section{Experiments on Link Prediction}
\label{sec:experimentation}
This section provides an empirical evaluation of the methods discussed in the previous section under a unified environmental settings and discusses the results based on the performance of the approaches applied to the task of link prediction. In this work, link prediction is chosen because most of the KG embedding models with literals are trained and evaluated on it. One of the major issues encountered while conducting these experiments is that the source code of some of these models is not openly available and is not easily reproducible. Such methods were excluded from the experimentation. In the subsequent sections, the datasets and the experiments with text, numeric, images and multi-modal literals are presented.

Based on the results of the experiments, a clear comparison is presented between the models with literals on link prediction. In addition, these models are also compared with the standard KG embedding approaches that they extend. Note that these models may inherit the problems that already exist in their corresponding base models - the standard KG embedding models that they extend). For instance, the models that extend DistMult such as DistMult-LiteralE$_{g}$ inherit the problem of DistMult, which is not being capable to properly capture anti-symmetric relations due to the way its scoring function is defined.

\subsection{Datasets}

The performance of the aforementioned models was measured using two of the most commonly used datasets for link prediction, i.e., FB15K~\cite{bordes2013translating} and FB15K-237~\cite{toutanova-chen-2015-observed} are considered. FB15K is a subset of Freebase \cite{bollacker2008freebase} which mostly contains triples describing the facts about movies, actors, awards, sports and sport teams. It contains a randomly split training, validation, and test sets. The issue with this dataset is that the test set contains a large number of triples which are obtained by simply inverting triples in the training set. This enables a simple embedding model which is symmetric with respect to the head and tail entity to obtain an excellent performance. In order to avoid this, the dataset FB15K-237 has been created by removing the inverse relations from FB15K. The statistics of these datasets is given in Table~\ref{table:datasets}.

\begin{table}[h]
\centering 
\caption{The number of entities, object relations, data relations, relational triples, 
train sets, valid sets, and test sets of the FB15K and the FB15K-237 datasets.}
\label{table:datasets}
\begin{tabular}{|c|c|c|}
\hline
& \multicolumn{2}{|c|}{\textbf{Datasets}}\\
\cline{2-3}
&\textbf{FB15K} &  \textbf{FB15K-237} \\
\hline
Entities & 14951 & 14541 \\ \hline
Object Relations & 1345 & 237\\ \hline
Relational triples& 592213 & 310116\\ 
\hline
Train sets& 483142 & 272115\\ \hline
Valid sets& 50000 & 17535\\ \hline
Test sets& 59071 & 20466\\ \hline
\end{tabular}
\end{table}

\subsection{Experiments with Text Literals} \label{sub:exp-text} 
As discussed in Section~\ref{text}, the embedding models Extended RESCAL, DKRL, KDCoE, and KGloVe with literals utilize text literals. However, all of these models except DKRL are not considered for experimentation due to the following issues:

\begin{itemize}
    \item The implementation of the model KGloVe with literals is not publicly available and it is not easily reproducible.
    
    \item KDCoE is designed specifically for cross-lingual entity alignment task which makes it difficult to apply it for link prediction.
    
    \item In case of Extended RESCAL, practically this method is computationally expensive and thus not considered as a feasible embedding model to incorporate literals. 
    
    Moreover, none of the models with literals which are discussed in this paper consider Extended RESCAL in their experiments. 
    
\end{itemize}  

In  order  to  conduct  experiments with text literals, 15239 English entity descriptions of the entities common in both datasets FB15K and FB15K-237 shown in Table~{\ref{table:datasets}} are taken from LiteralE {\cite{LiteralE2019}}. The focus lies on the common entity descriptions, i.e., for those entities existing in FB15K but not in FB15K-237 no description is used, because there has already been experiments done using the whole entity descriptions for FB15K dataset in the original paper. This way it would be possible to analyse the effect of the size of the dataset (the entity descriptions) on the performance of the embedding models. The average number of words (tokens) in the descriptions is 143 whereas the maximum and minimum are 804 and 2.

\paragraph{\textbf{
Dataset Pre-processing:}}
For pre-processing of the text (the entity descriptions), spacy.io\footnote{\url{https://spacy.io/usage}} has been used. This includes tokenization, named entity recognition and conversion of numbers to text, i.e., 16 has been converted to `sixteen'.  After the pre-processing step, all the entities along with the corresponding triples having no or short description of less than 3 words are removed. Also, the triples containing these entities are removed as mentioned by the authors in the paper. Moreover, only one description is chosen randomly for the entities with multiple text descriptions.

\paragraph{\textbf{Experimental Setup:}}
The hyperparameters used for DKRL are as follows: learning rate 0.001, embedding size 100, loss margin 1, batch size 100 and epochs 1000. For TransE, learning rate 0.01, embedding size 50, margin 1, and epochs 1000 are used. The experiments with DKRL were performed on Ubuntu 16.04.5 LTS system with 503GiB RAM and 2.60GHz speed. On the other hand,the experiments with TransE are performed with TITAN X (Pascal) GPU. 

\paragraph{\textbf{
Runtime:}}
Note that the codes used in the experiments for both models DKRL and TransE are not implemented in the same environment, i.e., for DKRL, the code that is released by the authors of the paper is used and for TransE the code provided by the authors of TransEA {\cite{Wu2018KnowledgeGE}} is used. Therefore, it is not fair to compare the runtime results of these two models directly. However, in order to provide some insights into the computational complexity of the models, their runtime results on the FB15K dataset are given as follows. DKRL takes 142 seconds to train 1 epoch using 16 threads whereas the runtime of TransE for a single iteration with batch size 4831 is 3.271 millisecond. This is computed by taking the average of 1000 iterations.  

\paragraph{\textbf{Evaluation Procedure and Results:}}
The performance of the model is evaluated based on the link prediction task. For each triple 
in the test set, a set of corrupted triples is generated with respect to the head or the tail entity. A triple is said to be corrupted with respect to its head entity if that head entity is replaced with any other entity from the KG, and analogously for a triple corrupted with respect to its tail entity. The set of corrupted triples can also contain true triples that exist in the training, validation or test set. Since it is not a mistake to give these true triples better score than the actual test triple, they are removed from the set of corrupted triples and this is referred to as filtered setting {\cite{bordes2013translating}}. In order to check if the model assigns a better score to the actual test triple than the corrupted triples which are obtained by corrupting the test triple, it is evaluated using the metrics MR (Mean Rank), MRR (Mean Reciprocal Rank), and Hits@N. First, for every test triple, all of its corrupted triples with respect to head are ranked based on their scores which are computed by the model. Then, the rank of the actual (true) test triples are taken in order to compute the metrics MR, MRR, and Hits@N. MR is the mean of the ranks of all test triples - the lower the better and MRR is their average inverse rank - the higher the better. Hits@N is the percentage of ranks lower than or equal to N - the higher the better. The same procedure is repeated to evaluate the model against the corrupted triples with respect to tail.

The results of link prediction on FB15K and FB15K-237 datasets are shown in Table~\ref{table:DKRL_table} for the models TransE, DKRL with Bernoulli distribution (DKRL$_{Bern}$), and DKRL with Uniform distribution (DKRL$_{unif}$). The Bernoulli distribution for sampling as defined in \cite{TransH} is a probability distribution, \(\frac{tph}{tph+hpt}\), where $tph$ is the average number of tail entities per head entity and $hpt$ is the average number of head entities per tail entity. Given a golden triple $<h, r, t>$, with the aforementioned probability, the triple is corrupted by replacing the head, and with probability \(\frac{hpt}{tph+hpt}\), the triple is corrupted by replacing the tail. The results are reported separately for the head entity and tail entity along with the overall results obtained by taking the mean of the head and tail predictions. The best scores are the ones which are highlighted in bold text. 
The result of the TransE model is presented in order to allow a clear comparison with DKRL because, as shown in Table {\ref{table:extensions}}, DKRL extends TransE. This comparison would help to further analyse the advantages of using text literals for KG embedding. 

Note that in the original paper
, the result of DKRL on FB15K is slightly better than TransE. However, in our experiments, as the results in Table {\ref{table:DKRL_table}} indicate, on the FB15K dataset TransE achieves better result than both versions of DKRL on all metrics except MRR and MR. The reason for this is that, as mentioned above, the set of entity descriptions used in our experiments are common for both datasets FB15K and FB15K-237, i.e., there is less entity descriptions in our experiment than there is in the original paper for FB15K. This indicates that the size of the dataset (the entity description has impact on the performance of the model). On the other hand, on the dataset FB15K-237 TransE is outperformed by DKRL$_{Unif}$ with respect to MR and by DKRL$_{Bern}$ with respect to the rest of the metrics.

Furthermore, the result shows that DKRL model with Bernoulli distribution (DKRL$_{Bern}$) has better performance than the model with Uniform distribution (DKRL$_{unif}$) for both the datasets.  DKRL$_{Bern}$ works best for the prediction of head, relation, and tail with respect to MRR, Hits@1, and Hits@3 whereas the \(DKRL_{Unif}\) method works better according to MR for both the datasets. DKRL$_{Bern}$ works slightly better than \(DKRL_{Unif}\) for FB15K-237 dataset. 
It is to be noted that DKRL has better improvement over TransE on FB15K-237 as compared to FB15K dataset because the former one does not contain symmetric relations, i.e., incorporating textual data to a clean dataset, such as FB15K-237, allows capturing more semantics.

\begin{table*}[h]
    \caption{Experiment results using DKRL model on FB15K and FB15K-237 datasets.}
    \centering
    \begin{tabular}{lcccccc}
    \hline
    \multicolumn{7}{c}{\textbf{FB15K}} \\
    \hline
        & & MR & MRR & Hits@1 & Hits@3 & Hits@10 \\
        & Head & 142 & 0.219 & \textbf{0.241} & \textbf{0.447} & \textbf{0.622} \\
        & Tail & 109 & 0.249 & \textbf{0.339} & \textbf{0.514}  & \textbf{0.690} \\
        \multirow{-4}{*}{TransE} & All & 125 & 0.234 & \textbf{0.290} & \textbf{0.480} &  \textbf{0.656} \\
    \hline
        & Head & 162 & \textbf{0.289} & 0.179 & 0.336 & 0.502 \\
        & Tail & 122 & \textbf{0.356} & 0.24 & 0.408 & 0.577 \\
        \multirow{-3}{*}{DKRL$_{Bern}$} &All & 142 & \textbf{0.322} & 0.209 & 0.372 & 0.539 \\
         
        \hline
    \multirow{3}{*}{DKRL$_{Unif}$}&Head & \textbf{96} & \textbf{0.289} & 0.172 & 0.335 & 0.52 \\
        & Tail & \textbf{75} & 0.333 & 0.211 & 0.383 & 0.576\\
        &All & \textbf{85} &  0.311 & 0.191 & 0.359 & 0.548 \\
    \hline
        \multicolumn{7}{c}{\textbf{FB15K-237}} \\
    \hline
         & & MR & MRR & Hits@1 & Hits@3 
         & Hits@10 \\
         & Head & 468 & 0.094 & 0.081 & 0.163 & 0.287 \\
         & Tail & 255 & 0.190 & 0.233 & 0.373 & 0.517 \\
         \multirow{-4}{*}{TransE} & All & 361 & 0.142 & 0.157 & 0.268 & 0.402 \\ 
        \hline
         \multirow{3}{*}{DKRL$_{Bern}$}& Head &145 & \bf{0.294} & \bf{0.184} & \bf{0.337}  & \textbf{0.507} \\
        & Tail & 98 & \bf{0.359} & \bf{0.244} & \bf{0.410} & \bf{0.585} \\
        & All & 122 & \bf{0.327} & \bf{0.214} & \bf{0.374} & \textbf{0.546} \\
        \hline
        \multirow{4}{*}{DKRL$_{Unif}$} & Head & \textbf{104} & 0.275 & 0.166 &0.312& 0.494\\
        & Tail & \textbf{77} & 0.322 & 0.209 & 0.363 & 0.552\\
        & All & \textbf{91} & 0.298 & 0.187 & 0.337 &  0.523\\
    \hline
    \end{tabular}
    \label{table:DKRL_table}
    
\end{table*}

\subsection{Experiment with Numeric Literals} \label{sub:exp-numeric}
MT-KGNN, KBLRN, LiteralE, and TransEA are the KG embedding models which make use of numeric literals (see Section~\ref{numeric}). KBLN, the submodel of KBLRN, which excludes the relational information provided by graph feature methods is used in the experiment instead of the main model KBLRN. This is the case because KBLN is directly comparable with the other three models (i.e., MT-KGNN, LiteralE, and TransEA) whereas KBLRN is not. The code\footnote{\url{https://github.com/kk0spence/TransEA}} for the TransEA model is the original implementation from TransEA \cite{Wu2018KnowledgeGE} where as the source codes\footnote{\url{https://github.com/SmartDataAnalytics/LiteralE}} for the models MT-KGNN, KBLN, and LiteralE are taken from the implementation in LiteralE \cite{LiteralE2019}. As described in Section.~\ref{numeric}, the structure-based embedding component of MT-KGNN is based on a neural network and it is referred to as RelNet. However, in the version implemented in LiteralE \cite{LiteralE2019}, they have replaced RelNet with DistMult as a baseline in order to have a directly comparable MTKGNN-like method to their proposed approach. Thus, in this survey, the MT-KGNN-like model has been used instead of the original MT-KGNN model. 

Moreover, the model LiteralE has different varieties depending on the baseline model and the transformation function used. As discussed in Section~\ref{sec:techniques}, in LiteralE there are two transformation functions: $g$ (GRU based function) and $lin$ (a simple linear function), and there are three baseline models - DistMult, ConvE and ComplEx. Thus, in this experiment, six varieties of the LiteralE model are considered: DistMult-Literale$_{g}$, ComplEx-Literale$_{g}$, ConvE-Literale$_{lin}$, DistMult-Literale$_{lin}$, ComplEx-Literale$_{lin}$, and ConvE-Literale$_{lin}$. The datasets, the experimental setup, and the evaluation results are discussed in the subsequent sections.

\paragraph{\textbf{
Attributive Triples:}}
In order to conduct the experiments with numeric literals, both the datasets FB15K and FB15K-237 given in Table~\ref{table:datasets} are extended with a set of 23521 attributive triples, containing only numeric literals, with 118 data relations. These triples are created based on the attributive triples from TransEA \cite{Wu2018KnowledgeGE}. In TransEA, the authors have provided a set of attributive triples where the object values are numeric. However, it is not possible to directly use this data as the literal values are normalized in the interval [0-1] as required by the model but the other models in this experiment, like LiteralE, use the original unnormalized literal values instead. Therefore, it was necessary to query Freebase to replace the normalized object literal value for each (subject, data relation) pair from the TransEA attributive triples data. Moreover, only those data relations which occur in at least 5 triples are taken into consideration. 

\paragraph{\textbf{Experimental Setup:}}
For both datasets, the hyperparameters for TransEA are: epoch 3000, dimension 100, batches 100, margin 2, and learning rate 0.3 and for TransE they are described in Section~\ref{sub:exp-text}. 
For the other models, same as in LiteralE, the hyperparameters used for both datasets are: learning rate 0.001, batch size 128, embedding size 100 (for DistMult, ComplEx and their extensions with literals) and 200 (for KBLN, and MTKGNN, ConvE, and ConvE's extensions), embedding dropout probability 0.2, label smoothing 0.1, and epochs 1000 for ConvE and 100 for the rest. TITAN X (Pascal) GPU has been used for the models LiteralE, KBLN, and MTKGNN. 

\paragraph{\textbf{
Runtime:}}
As in the experiments with text literals, not all the models in the experiments with numeric literals are implemented in the same environment, i.e., for TransEA the code that is released by the authors of the paper is used and for the other models the code that is provided by the authors of LiteralE are used. Therefore, direct comparison of the runtime of TransEA and the other models would not be possible. However, the runtime of each of the models is computed on FB15K dataset so as to give insights into the models computational complexity. The running time of TransEA is 3.271 ms per a single iteration with batch size of 4831. For the other models their runtime for a single iteration of batch size 128 is shown in Table {\ref{tab:runtime}}. Note that the runtime results reported here are the average over runtime values of 1000 iterations.
\begin{table}[]
    \centering
    \caption{
    Runtime of models considered in the experiments with numeric literals. The resutls are per single iteration and reported in milliseconds.} 
    \begin{tabular}{c|c}
    \hline
        \textbf{
        } & \textbf{
        Time(ms)}
        \\
    \hline
        DistMult-LiteralE$_{g_{lin}}$ & 31.575  \\
        DistMult-LiteralE$_{g}$ & 37.138 \\
        ComplEx-LiteralE$_{g_{lin}}$ & 39.269\\
        ComplEx-LiteralE$_{g}$ & 52.346 \\
        ConvE-LiteralE$_{g_{lin}}$ &  43.386 \\
        ConvE-LiteralE$_{g}$ & 50.439\\
        KBLN &  86.825\\
        \hline
        DistMult  & 29.679 \\
        ComplEx &  33.526\\
        ConvE & 40.970\\
        \hline
    \end{tabular}
    \label{tab:runtime}
\end{table}
\paragraph{\textbf{Evaluation Procedure and Results:}} The same evaluation metrics which are discussed in Section \ref{sub:exp-text} has been used to evaluate the performance of the models with numeric literals on the link prediction task. As shown in Table~\ref{table:Fb15k-numeric}, according to the overall result, the model KBLN has considerably better performance than the other models in all metrics except MR. The results from the ComplEx-LiteralE$_g$ model 
show 
that it is capable to produce a highly competitive performance having the second best results with respect to the same metrics. This is the case due to the fact that this model is able to handle the inverse relations in FB15K by applying the complex conjugate of an entity embedding when the entity is used as a tail and its normal embedding when it is the head. 
Moreover, the model KBLN also achieves better result when compared to the standard models without literals presented in Table~{\ref{table:Fb15k-standard}} with respect to all metrics except MR.

Another possible analysis to make is to compare the results of the standard models presented in Table~{\ref{table:Fb15k-standard}} with the results of their extensions shown in the `both head and tail Prediction' part of Table~{\ref{table:Fb15k-numeric}}. For instance, ComplEx-LiteralE$_g$ achieves better performance than its base model ComplEx according to all metrics which indicates that using numeric literals with ComplEx by applying the approach in LiteralE is beneficial. However, this is not the case with DistMult and ConvE. One reason for this can be the fact that the number of attributive triples used in our experiment is not as big as in the original paper of LiteralE, i.e, increasing the number of numeric literals may improve the result as already seen in the original paper of literalE.

On the other hand, referring to the overall result on FB15K-237 dataset as shown in Table {\ref{table:Fb15k-237-numeric}}, the model DistMult-LiteralE$_g$ outperforms the other models according to all metrics. This entails that applying LiteralE to DistMult on FB15K-237 provides better performance than applying it to other baseline models. Note that the reason for DistMult-LiteralE$_g$ model to achieve the best result on FB15K-237 dataset, as comapred to  FB15K, may be due to the fact that this dataset does not have any symmetric relation, i.e., DistMult already has difficulties in modeling
asymmetric relations on FB15k and adding literals might introduce noise but in case of FB15K-237, incorporating literals improves DistMult because symmetric relations are removed. 
Regarding the two transformation functions $g$ and $g_{lin}$, the function $g$ leads to better results than $g_{lin}$ according to the results on both dataset.

\begin{table*}[h]
\centering 
\caption{Link prediction results on FB15K dataset using filtered setting.}
\label{table:Fb15k-numeric}
\resizebox{8.25cm}{!}{
\begin{tabular}{lccccc}
\hline
\multicolumn{6}{c}{\textbf{Head Prediction}}\\
\hline
Models & MR & MRR & Hits@1 & Hits@3 & Hits@10 \\ \hline
DistMult-LiteralE$_{g_{lin}}$ & 121 & 0.495 & 0.383 & 0.559 &  0.697 \\
ComplEx-LiteralE$_{g_{lin}}$ & 71 & 0.76 & 0.697 &  0.801 &  0.876 \\
ConvE-LiteralE$_{g_{lin}}$ & 52 & 0.612 &  0.51 & 0.678 &  0.795 \\
DistMult-LiteralE$_{g}$ & 72 & 0.581 & 0.479 & 0.642 & 0.762 \\
ComplEx-LiteralE$_{g}$ & 63 & 0.768 & \textbf{0.707} & 0.809 & 0.878 \\
ConvE-LiteralE$_{g}$ & \textbf{49} & 0.72 & 0.65 & 0.762 & 0.849 \\
KBLN & 77 & \textbf{0.775} & 0.705 & \textbf{0.827} & \textbf{0.892} \\
MTKGNN  & 73 & 0.702 & 0.617 & 0.758 & 0.855 \\
TransEA  & 103 & 0.285 & 0.367 & 0.609 & 0.728 \\
 \hline
\multicolumn{6}{c}{\textbf{Tail Prediction}}\\ \hline
Models & MR & MRR & Hits@1 & Hits@3 & Hits@10 \\
\hline
DistMult-LiteralE$_{g_{lin}}$ & 145 & 0.447 & 0.337 & 0.507 & 0.645 \\
ComplEx-LiteralE$_{g_{lin}}$  & 101 & 0.704 & 0.64 & 0.743 & 0.821 \\
ConvE-LiteralE$_{g_{lin}}$   & \textbf{74} & 0.567 & 0.465 & 0.63 & 0.746 \\
DistMult-LiteralE$_{g}$ & 94 & 0.528 & 0.425 & 0.589 & 0.712 \\
ComplEx-LiteralE$_{g}$ & 93 & 0.711 & 0.65 & 0.746 & 0.821 \\
ConvE-LiteralE$_{g}$ & 79 & 0.657 & 0.586 & 0.698 & 0.783 \\
KBLN & 90 & \textbf{0.727} & \textbf{0.656} & \textbf{0.776} & \textbf{0.848} \\
MTKGNN  & 91 & 0.65 & 0.562 & 0.708 & 0.806 \\
TransEA  & 75 & 0.314 & 0.417 & 0.671 & 0.805 \\
\hline
\multicolumn{6}{c}{\textbf{Both Head and Tail Prediction}}\\
\hline
Models & MR & MRR & Hits@1 & Hits@3 & Hits@10 \\ \hline
DistMult-LiteralE$_{g_{lin}}$  & 133 & 0.471 & 0.36 & 0.533 & 0.671 \\
ComplEx-LiteralE$_{g_{lin}}$  & 86 & 0.732 & 0.668 & 0.772 & 0.848 \\
ConvE-LiteralE$_{g_{lin}}$ & \textbf{63} & 0.589 & 0.487 & 0.654 & 0.77 \\
DistMult-LiteralE$_{g}$ & 83 & 0.554 & 0.452 & 0.615 & 0.737 \\
ComplEx-LiteralE$_{g}$ & 78 & 0.739 & 0.678 & 0.777 & 0.849 \\
ConvE-LiteralE$_{g}$ & 64 & 0.688 & 0.618 & 0.73 & 0.816 \\
KBLN & 83 & \textbf{0.751} & \textbf{0.68} & \textbf{0.801} & \textbf{0.87} \\
MTKGNN  & 82 & 0.676 & 0.589 & 0.733 & 0.83 \\
TransEA  & 74 & 0.299 & 0.392 & 0.64 & 0.766 \\ \hline
\end{tabular}
}
\end{table*}

\begin{table*}[h]
\centering 
\caption{
Link prediction results with models without literals on FB15K using filtered setting.}
\label{table:Fb15k-standard}
\resizebox{8.25cm}{!}{
\begin{tabular}{p{2.5cm}ccccc}
\hline
\multicolumn{6}{c}{\textbf{FB15K}}\\
\hline
Models & MR & MRR & Hits@1 & Hits@3 & Hits@10 \\ \hline
DistMult & 119 & 0.67 & 0.589 & 0.723 & 0.817 \\
ComplEx  & 127 & \textbf{0.692} & \textbf{0.614} & 0.742 & 0.833 \\
ConvE & \textbf{50} & 0.689 & 0.593 & \textbf{0.757} & \textbf{0.852} \\
TransE & 125 & 0.234 & 0.290 & 0.480 &  0.656 \\
\hline
\end{tabular}
}
\end{table*}

\begin{table*}[h!]
\centering 
\caption{Link prediction results on FB15K-237 dataset using filtered setting.}
\label{table:Fb15k-237-numeric}
\resizebox{8.25cm}{!}{
\begin{tabular}{lccccc}
\hline
\multicolumn{6}{c}{\textbf{Head Prediction}}\\ \hline
Models & MR & MRR & Hits@1 & Hits@3 & Hits@10 \\ \hline
DistMult-LiteralE$_{g_{lin}}$ & 245 & 0.377 & 0.279 & 0.422 &  0.568 \\
ComplEx-LiteralE$_{g_{lin}}$ & 371 & 0.36 & 0.271 &  0.4 &  0.538 \\
ConvE-LiteralE$_{g_{lin}}$ & \textbf{208} & 0.388 &  0.296 & 0.427 &  0.572 \\
DistMult-LiteralE$_{g}$ & 209 & \textbf{0.413} & \textbf{0.320} & \textbf{0.456} & \textbf{0.591} \\
ComplEx-LiteralE$_{g}$ & 315 & 0.366 & 0.277 & 0.404 & 0.543 \\
ConvE-LiteralE$_{g}$ & 236 & 0.317 & 0.229 & 0.345 & 0.501 \\
KBLN & 381 & 0.386 & 0.295 & 0.426 & 0.564 \\
MTKGNN  & 437 & 0.383 & 0.295 & 0.423 & 0.559 \\
TransEA  & 389 & 0.111 & 0.094 & 0.197 & 0.342 \\
\hline
\multicolumn{6}{c}{\textbf{Tail Prediction}}\\ \hline
Models & MR & MRR & Hits@1 & Hits@3 & Hits@10 \\ \hline
DistMult-LiteralE$_{g_{lin}}$ & 426 & 0.195 & 0.119 & 0.214 & 0.349 \\
ComplEx-LiteralE$_{g_{lin}}$ & 575 & 0.17 & 0.104 & 0.185 & 0.306 \\
ConvE-LiteralE$_{g_{lin}}$  & 362 & 0.187 & 0.112 & 0.204 & 0.338 \\
DistMult-LiteralE$_{g}$ & 359 & \textbf{0.215} & 0.137 & 0.234 & 0.371 \\
ComplEx-LiteralE$_{g}$ & 493 & 0.175 & 0.106 & 0.19 & 0.312 \\
ConvE-LiteralE$_{g}$ & 459 & 0.131 & 0.07 & 0.137 & 0.256 \\
KBLN & 501 & 0.207 & 0.128 & 0.23 & 0.362 \\
MTKGNN & 580 & 0.191 & 0.12 & 0.208 & 0.338 \\
TransEA & \textbf{203} & 0.206 & \textbf{0.25} &\textbf{ 0.409} & 0.57 \\ \hline
\multicolumn{6}{c}{\textbf{Both Head and Tail Prediction}} \\ \hline
Models & MR & MRR & Hits@1 & Hits@3 & Hits@10 \\ \hline
DistMult-LiteralE$_{g_{lin}}$  & 335
 & 0.286 & 0.199 & 0.318 & 0.458 \\
ComplEx-LiteralE$_{g_{lin}}$  & 473 & 0.265 & 0.187 & 0.292 & 0.422 \\
ConvE-LiteralE$_{g_{lin}}$  & 285 & 0.287 & 0.204 & 0.315 & 0.455 \\
DistMult-LiteralE$_{g}$ & \textbf{284} & \textbf{0.314} & \textbf{0.228} & \textbf{0.345} & \textbf{0.481 }\\
ComplEx-LiteralE$_{g}$ & 404 & 0.27 & 0.191 & 0.297 & 0.427 \\
ConvE-LiteralE$_{g}$ & 347 & 0.224 & 0.149 & 0.241 & 0.378 \\
KBLN & 441 & 0.296 & 0.211 & 0.328
 & 0.463 \\
MTKGNN  & 508 & 0.287 & 0.207 & 0.315 & 0.448 \\
TransEA  & 296 & 0.158 & 0.172 & 0.303 & 0.456 \\ \hline
\end{tabular}
}
\end{table*}

\begin{table*}[h]
\centering 
\caption{
Link prediction results with models without literals on FB15K-237 dataset using filtered setting.}
\label{table:Fb15k-237-standard}
\resizebox{8.25cm}{!}{
\begin{tabular}{p{2.5cm}ccccc}
\hline
Models & MR & MRR & Hits@1 & Hits@3 & Hits@10 \\ \hline
DistMult & 630 & 0.280 & 0.201 & 0.309 & 0.438 \\
ComplEx  & 623 & 0.288 & 0.207 & 0.318 & 0.448 \\
ConvE & \textbf{273} & \textbf{0.310} & \textbf{0.222} & \textbf{0.343} & \textbf{0.484} \\

TransE & 361 & 0.142 & 0.157 & 0.268 & 0.402 \\ \hline
\end{tabular}
}
\end{table*}

\subsection{Experiment with Images} \label{sub:exp-images} 
Note that it is not possible to compare the whole of MKBE \cite{mmkb:emnlp18} with any other model as it is the only embedding model which utilizes the three types of literals together: text, numeric, and images. Therefore, its sub model S+I which uses only images has been compared with the embedding model IKRL \cite{Xie2017ImageembodiedKR}. Since this comparison has already been done by the authors of MKBE \cite{mmkb:emnlp18}, the result shown in Table \ref{table:results-image} is directly taken from their paper. They have compared  the models DistMult+S+I, ConvE+S+I, and IKRL where S stands for structure and I for Image. Both DistMult+S+I and ConvE+S+I are sub models of MKBE which use only relational triples and Images. The result indicates that ConvE+S+I outperforms the other two models in all metrics on the YAGO-10 dataset (refer to MKBE \cite{mmkb:emnlp18} for more details).

\begin{table*}[ht]
\centering 
\caption{MRR results on link prediction task on YAGO-10 taken from MKBE
\cite{mmkb:emnlp18}.}
\label{table:results-image}
\begin{tabular}{ccccc}
\hline
\multicolumn{5}{c}{\textbf{YAGO-10}}\\
Models & MRR & Hits@1 & Hits@3 & Hits@10 \\
\hline
DistMult+S+I & 0.342 & 0.235 & 0.352 & 0.618 \\
ConvE+S+I &   \textbf{0.566} & \textbf{0.471} & \textbf{0.597} & \textbf{0.72} \\
IKRL &   0.509 & 0.423 & 0.556 & 0.663 \\
\hline
\end{tabular}
\label{fig:figures}
\end{table*}

\subsection{Experiment with Multi-modal Literals} \label{sub:exp-multimodal} 
As discussed in Section~\ref{sec:techniques}, the existing multi-modal embeddings are categorized into two types: i) models with text literal, numeric literal and image literals and ii) models with text and numeric literals. However, since MKBE is the only model in the first category only its submodel $S+I$ could be compared with IKRL (see Section \ref{sub:exp-images}). Regarding the models with text and numeric literals, i.e., LiteralE with blocking and EAKGAE, they are not included in the experiment as well. The issue with EAKGAE is the same as that of KDCoE, i.e., it is trained on entity alignment task where as the reason for not having LiteralE with blocking is that its code is not publicly available. On the contrary, LiteralE (a model with numerics) has also been adopted to incorporate text literals in the experiments conducted by the authors. Similarly, in our experiment, the LiteralE approach has been tried out with the combination of text and numeric literals, i.e., the model DistMult-LiteralE$_g$-text in Table \ref{tab:text-and-numerics}. Then, the result has been compared with LiteralE with just numeric literals (DistMult-LiteralE$_g$) and DKRL (a model using only text literals) so as to investigate the benefits of utilizing information represented by different types of literals. 

DistMult-LiteralE$_g$-text is a model which applies the LiteralE approach to DistMult by using both numeric and text literals. Note that DistMult is chosen here as a baseline due to the reason that the best result in the experiments with numerics on the FB15K-237 dataset is achieved using this model as discussed in Section~\ref{sub:exp-numeric}. The  datasets listed in Table~\ref{table:datasets} are also used for this experiment along with additional text attributive triples which are descriptions of entities. DistMult-LiteralE$_g$-text has also been compared with its numeric only equivalent DistMult-LiteralE$_g$ and DKRL$_{Bern}$. 

The experimental results obtained on the datasets FB15K and FB15K-237 are shown in Table~\ref{tab:text-and-numerics}. As the result indicates, combining text and numeric literals on FB15K dataset with DistMult-LiteralE$_g$-text approach does not produce better results as compared to the other models DistMult-LiteralE$_g$ and DKRL$_{Bern}$.  As mentioned before, this dataset contains a set of inverse relations which may lead to having a triple whose inverse has a different label. Given the fact that DistMult fails to model such asymmetric relations, incorporating more literals with DistMult may introduce much noise than improving the performance. On the other hand, for FB15K-237 dataset, according to all the measures except MR, DistMult-LiteralE$_g$-text model works better for the head entity prediction compared to the other two models. For tail entity prediction, DKRL$_{Bern}$ works better with respect to all measures for the same dataset. 
\begin{table*}[h]
    \centering
    \caption{Link prediction results on FB15K and FB15K-237 datasets using filtered set.}
    \label{tab:text-and-numerics}
    \begin{tabular}{lcccccc}
     \hline
        \multicolumn{7}{c}{\textbf{FB15K}}\\  \hline
        & Models & MR & MRR & Hits@1 & Hits@3 & Hits@10 \\ 
        \hline
        \multirow{2}{*}{Head} & \pbox{7cm}{DistMult-LiteralE$_g$} &  \textbf{72} & \textbf{0.581} & \textbf{0.479} & \textbf{0.642} & \textbf{0.762} \\
        & DKRL$_{Bern}$ & 162 & 0.289 & 0.179 & 0.336 & 0.502 \\
        & \pbox{7cm}{DistMult-LiteralE$_g$-text} & 93 & 0.516 & 0.405 & 0.582 & 0.711 \\ \cline{2-7}
        \multirow{2}{*}{Tail} & \pbox{7cm}{DistMult-LiteralE$_g$} &  \textbf{94} & \textbf{0.528} & \textbf{0.425} & \textbf{0.589} & \textbf{0.712} \\
        & DKRL$_{Bern}$ & 122 & 0.356 & 0.24 & 0.408 & 0.577 \\
        & \pbox{7cm}{DistMult-LiteralE$_g$-text} & 119 & 0.463 & 0.351 & 0.532 & 0.66 \\  \cline{2-7}
        \multirow{2}{*}{All} & \pbox{7cm}{DistMult-LiteralE$_g$} &  \textbf{83} & \textbf{0.554} & \textbf{0.452} & \textbf{0.615} & \textbf{0.737} \\
        & DKRL$_{Bern}$ & 142 & 0.322 & 0.209 & 0.372 & 0.539 \\
        & \pbox{7cm}{DistMult-LiteralE$_g$-text} & 106 & 0.489 & 0.378 & 0.557 & 0.685 \\  \hline
        \multicolumn{7}{c}{\textbf{FB15K-237}}\\  \hline
        & Models & MR & MRR & Hits@1 & Hits@3 & Hits@10 \\
        \hline
        \multirow{2}{*}{Head} & DistMult-LiteralE$_g$ & 209 & 0.413 & 0.320 & 0.456 & 0.591 \\
        & DKRL$_{Bern}$ & \textbf{145} & 0.294 & 0.184 & 0.337 & 0.507 \\
        & \pbox{7cm}{DistMult-LiteralE$_g$-text} & 207 & \textbf{0.416} & \textbf{0.323} & \textbf{0.462} & \textbf{0.594} \\  \cline{2-7}
        \multirow{2}{*}{Tail} & DistMult-LiteralE$_g$ & 359 & 0.215 & 0.137 & 0.234 & 0.371 \\
        & DKRL$_{Bern}$ & \textbf{98} & \textbf{0.359} & \textbf{0.244} & \textbf{0.410} & \textbf{0.585} \\
        & \pbox{7cm}{DistMult-LiteralE$_g$-text} & 354 & 0.223 & 0.142 & 0.246 & 0.385 \\ 
        \cline{2-7}
        \multirow{2}{*}{All} &
        DistMult-LiteralE$_g$ & 284 & 0.314 & 0.228 & 0.345 & 0.481 \\
        & DKRL$_{Bern}$ & \textbf{122} & \textbf{0.327} & 0.214 & \textbf{0.374} & \textbf{0.546} \\
        & \pbox{7cm}{DistMult-LiteralE$_g$-text} & 280 & 0.319 & \textbf{0.232} & 0.354 & 0.489 \\  \hline
    \end{tabular}
\end{table*}

\section{Discussion and Conclusion}
\label{sec:discussion}
Given the recent massive attention towards the use of KGs in various applications, different KG embedding techniques have been proposed to enable efficient use of KGs. 
In some of these techniques, an attempt has been made to utilize the information represented in literals present in KGs for a better quality embedding of the elements of the KGs, i.e., entities and relations. In this paper, a comprehensive survey of those KG embedding models with literals has been presented. The survey provides a detailed analysis and categorization of these models based on the proposed methodology along with their application scenarios and limitations. Moreover, various experiments on link prediction task on these models have been conducted so as to compare the models' performances. 

In this paper, two major research questions are formulated and presented in Section~{\ref{sec:problem_statement}}. The answers to these questions are given as follows: 
\begin{itemize}
\item \textbf{
RQ1} 
-- \textit{
How can structured (triples with object relations) and unstructured information (attributive triples) in the KGs be combined into the representation learning?} 
\\
In order to use both data sources, i.e., triples with object relations and triples with data relations (attributive triples) together for representation learning, in broader terms, the following two techniques are considered in the models discussed in this paper: 
\begin{itemize}
    \item 
    Handling literals separately: defining one task per data source like in TransEA or using a separate encoder for literals as in DKRL. The two tasks are trained simultaneously to make sure that for every entity the information available in both data sources are used to learn its embedding. The embeddings of the entities learned based on each data source can be unified in the vector space or not depending on how the model works. For instance, Jointly(Disp) learns unified representation for entities where as DKRL generates two representations per entity and do not force them to be unified. 
    \item 
    Incorporating literals directly into entity embeddings: as in LiteralE, one way is to extend a certain latent feature method by directly enriching the embeddings with information from literals via a learnable parameter and use the same scoring function from the latent feature method. 
\end{itemize}
\item \textbf{
RQ2} 
-- \textit{
How can the heterogeneity of the types of literals present in the KGs be captured and combined into representation learning?} 
\\
The following are some possible ways to combine different kind of literals, i.e., text, numeric, etc. together for representation learning. 
\begin{itemize}
    \item 
    Encoding each type of literal separately: in order to capture the semantics of literals, different encoders can be used for different types of literals, for example, CNN for textual descriptions. Then, as shown in MKBE, each attributive triple can be treated same as structured triples and use a single scoring function for training. 
    \item 
    Incorporating information present in every kind of literal directly into the entity embedding: as in LiteralE, for a given entity, first the literals associated with it are encoded as vectors - using one vector per type of literal. Then, a mapping function is used to map all these vectors (including the structure-based vector representation of the entity) into a single vector.
\end{itemize}
\end{itemize}
As mentioned in Section~\ref{sec:techniques} or seen from the result of the experiments in Section~\ref{sec:experimentation}, these embedding models have different drawbacks such as: 

\begin{itemize}
    \item The effect that data types/units have on the semantics of literals has not been considered by any of the models.
    \item Most of the embedding models which make use of numerical literals, such as LiteralE, TransEA, MT-KGNN, and KBLN consider only the year part of date typed literals and ignore the month and day values. This hinders the ability to properly capture the information represented in such kind of literals. For example, given the following three date typed literal values:
    \begin{lstlisting} 
    "1999-10-29"^^xsd:date,
    "1999-04-14"^^xsd:date, 
    "1999-10-30"^^xsd:date, 
    \end{lstlisting}
    a model which utilizes only the year part of these values considers all of them to be exactly the same despite the fact that the first date value is more close to the third value than it is to the second value. 
    \item Most of the models also do not have a proper mechanism to handle multi-valued literals. 
    \item The performance of most of the models is dependent on the dataset used for training and testing which shows that these models are not robust. 
    For example, referring to Table~\ref{tab:text-and-numerics}, the results of the model DistMult-LiteralE$_{g}$-text indicate that combining text and numeric literals yields better performance on FB15K-237 but not on FB15K due to the technique used in the model and the nature of the datasets (see Section~\ref{sub:exp-multimodal}).
    \item Not all the models are effective in combining different types of literals. 
    For example, the performance of DistMult-LiteralE$_{g}$-text (numeric + text literals), which combines text and numeric literals, on the dataset FB15K is lower as compared to DistMult-LiteralE$_{g}$ (only numeric literals). 
    \item Only few approaches have been proposed for multi-modal KG embeddings and none of them take into consideration literals with URIs connected to items such as audio, video, or pdf files. 
\end{itemize}
The above described shortcomings of the existing models clearly indicate that thorough investigation is needed on how to address different types of literals that obtain different inherent semantics. For instance, a possible perspective that arises by this detailed analysis is that there is a need to properly handle the data typed literals such as the values of the data relation \textit{weight} given in \textit{kilogram} and \textit{pound}. One possible solution to target this issue could be to normalize these literal values to a standardized measures and to treat different measures like weights and lengths separately in the representation learning process. 

One cannot expect that by leaving out available information present in the original KG, its latent representation as being only an approximation of the original KG, will perform equally well on tasks that depend on its semantic information content.
Overall, the inclusion of datatyped literals with a proper representation of their semantics into the representation learning process will increase the model's semantic content and might thereby lead to quality improvement.

\nocite{*} 
\bibliographystyle{ios1}  
\bibliography{biblio} 

\clearpage
\begin{landscape}
\appendix
\section{Summary of Applications}
\vspace*{\fill}
    \center %
    \captionof{table}{
    Summary of different applications on which the KG embedding techniques with literals, in their original papers, have been trained and/or evaluated} 
    \label{table:application-summary}
     \vspace{1cm}
    \bgroup
    \def\arraystretch{1.5} 
    \begin{tabular}{|l|c|c|c|c|c|c|c|c|
    c|}
     \hline
        & \pbox{2cm}{Link \\ Prediction} & \pbox{3cm}{Triple \\ Classification} & \pbox{3cm}{Entity \\ Classification} & \pbox{2cm}{Entity \\  Alignment} &
        \pbox{3cm}{Attribute \\ value \\prediction} &\pbox{3cm}{Nearest\\ neighbour\\ analysis} & \pbox{2cm}{Data \\ linking} &\pbox{2cm}{Document \\ classification} & \pbox{2cm}{Relational \\ Fact \\ Extraction}\\  \hline      
        Extended RESCAL  &\checkmark&&\ 
        \checkmark
        &&&&&&\\ \hline LiteralE &\checkmark&&&&&\checkmark&&&\\ \hline 
        TransEA &\checkmark&&&&&&&&\\ \hline
        KBLRN &\checkmark&&&&&&&&\\ \hline
        DKRL &\checkmark&&\checkmark&&&&&&\\ \hline
        KDCoE &\checkmark&&&\checkmark&&&&&\\ \hline
        KGlove with literals &&&&&&&&\checkmark&\\ \hline IKRL&\checkmark&\checkmark&&&&&&&\\ \hline
        EAKGE &\checkmark&&&\checkmark&&&&&\\ \hline
        MKBE &\checkmark&&&&\checkmark&&&&\\ \hline
        MT-KGNN &&\checkmark&&&\checkmark&&&&\\ \hline
        LiteralE with blocking &&&&&&&\checkmark&&\\ \hline
        Jointly(Desp) &\checkmark&\checkmark&&&&&&&\checkmark\\ \hline
        Jointly &\checkmark&\checkmark&&&&&&&\\ \hline
        SSP &\checkmark&&\checkmark&&&&&&\\ \hline
        MTKGRL &\checkmark&\checkmark&&&&&&&\\ 
         \hline
    \end{tabular}
    \egroup
    \vfill  
\end{landscape}
\end{document}